\documentclass[10pt,twocolumn,letterpaper]{article}

\usepackage{wacv}
\usepackage{times}
\usepackage{epsfig}
\usepackage{graphicx}
\usepackage{amsmath}
\usepackage{amssymb}
\usepackage{csquotes}
\usepackage{pbox}

\graphicspath{{fig/}}
\DeclareGraphicsExtensions{.ps,.eps,.epsi,.png,.pdf}

\wacvfinalcopy % *** Uncomment this line for the final submission

\ifwacvfinal\fi
\begin{document}

%%%%%%%%% TITLE
%\title{Weather Property Estimation for Images Using CNN and RNN}
\title{Visual Weather Temperature Prediction}
%\title{Weather Property Estimation using Convolutional Recurrent Neural Networks}

% Authors at the same institution
%\author{Wei-Ta Chu \hspace{2cm} Kai-Chia Ho \\
%National Chung Cheng University\\
%{\tt\small wtchu@ccu.edu.tw}
%}
% Authors at different institutions
\author{Wei-Ta Chu \\
National Chung Cheng University\\
{\tt\small wtchu@ccu.edu.tw}
\and
Kai-Chia Ho \\
National Chung Cheng University\\
{\tt\small kevinho627627@gmail.com}
\and
Ali Borji \\
Center for Research in Computer Vision, University of Central Florida\\
{\tt\small aborji@crcv.ucf.edu}
}

\maketitle
\ifwacvfinal\thispagestyle{empty}\fi

%%%%%%%%% ABSTRACT
\begin{abstract}
In this paper, we attempt to employ convolutional recurrent neural networks
for weather temperature estimation using only image data. We study ambient temperature estimation based on deep neural networks in two scenarios a) estimating temperature of a single outdoor image, and b) predicting temperature of the last image in an image sequence. In the first scenario, visual features are extracted by a convolutional neural network trained on a large-scale image dataset. We demonstrate that promising performance can be obtained, and analyze how volume of training data influences performance. In the second scenario, we consider the temporal evolution of visual appearance, and construct a recurrent neural network to predict the temperature of the last image in a given image sequence. We obtain better prediction accuracy compared to the state-of-the-art models. Further, we investigate how performance varies when information is extracted from different scene regions, and  when images are captured in different daytime hours. Our approach further reinforces the idea of using only visual information for cost efficient weather prediction in the future.
\end{abstract}

%%%%%%%%% BODY TEXT
\section{Introduction}
Visual attributes of images have been heavily studied for years. Most previous works have been focused on recognizing \enquote{explicit visual attributes} in images, such as object's texture and color distributions \cite{ferrari07}, and semantic categories \cite{jaya14}. With the advance of computer vision and machine learning technologies, more and more works have been proposed to study discovering \enquote{subtle attributes} in images. These subtle attributes may not be well-formulated in explicit forms, but are usually recognizable by human beings. For example, Lu et al.~\cite{lu17} proposed a method to recognize whether an image was captured on a sunny day or on a cloudy day. Hays and Efros \cite{hays08} proposed to estimate geographic information from a single image (a.k.a IM2GPS). Recent research has demonstrated that deep learning approaches are effective for recognizing painting styles~\cite{karayev14}\cite{chu16}. 
 
Among various subtle attributes, weather properties of images have attracted increasing attention. The earliest investigation of the relationship between vision and weather conditions dates back to early 2000s \cite{nara02}. Thanks to the development of more advanced visual analysis and deep learning methods, a new wave of works studying the correlation between visual appearance and ambient temperature or other weather properties has recently emerged~\cite{glasner15}\cite{zhou17}. Inspired by these works, here we mainly focus on ambient temperature prediction from a single image or a sequence of images. We will investigate different deep learning approaches and demonstrate that the state-of-the-art performance can be obtained.

\begin{figure*}[ht]
\centering
\includegraphics[height=2.05cm]{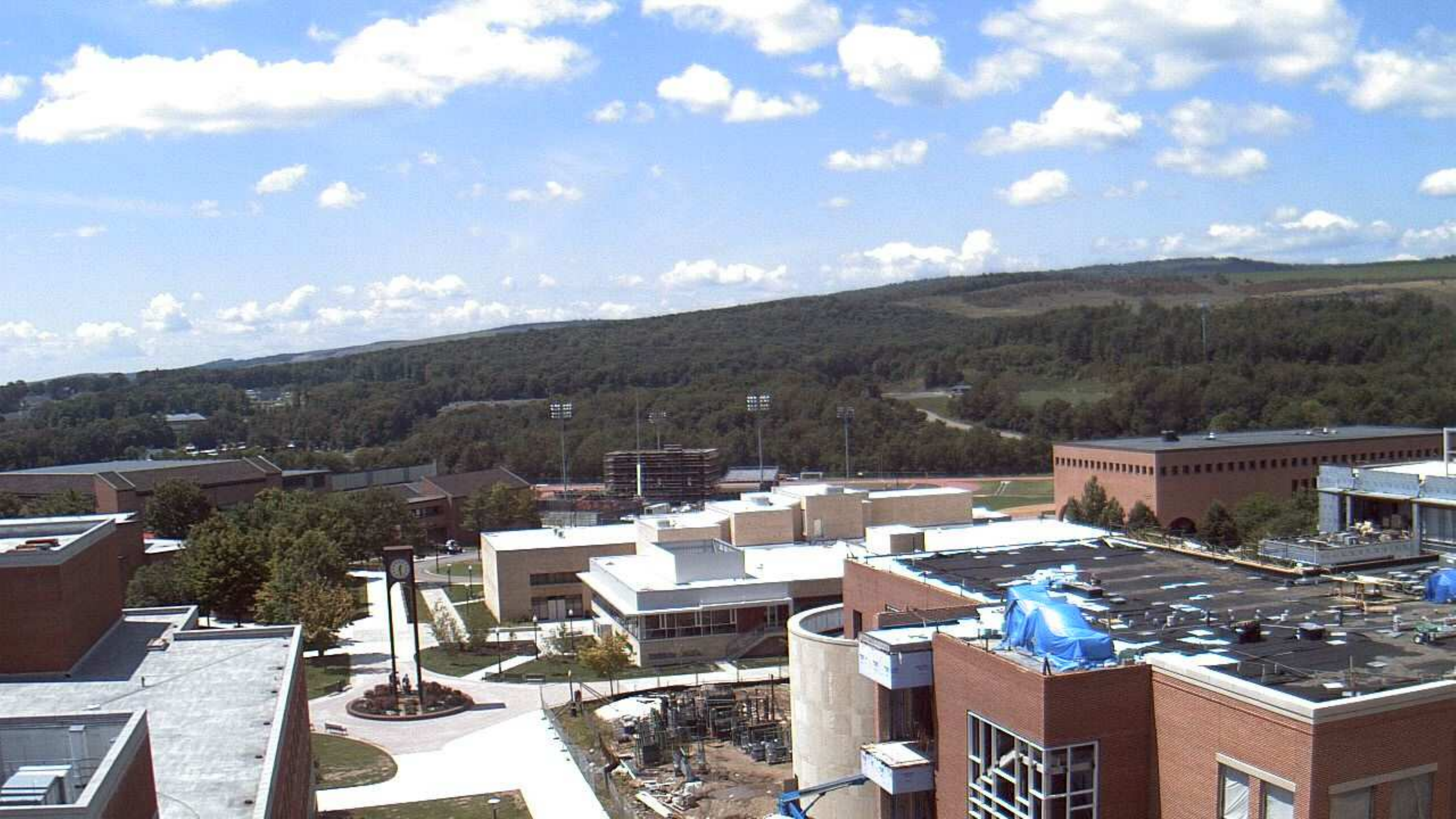} 
\includegraphics[height=2.05cm]{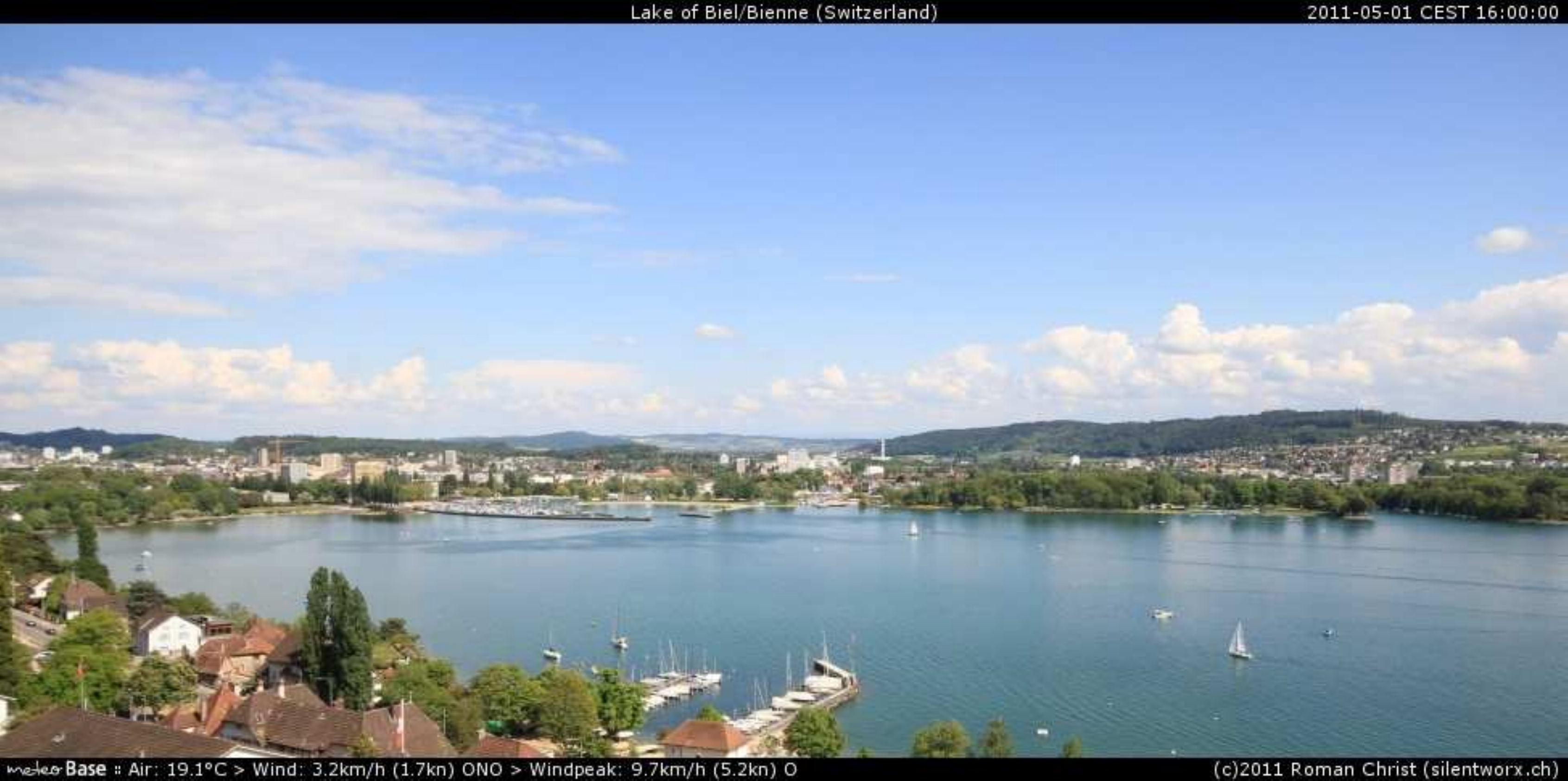} 
\includegraphics[height=2.05cm]{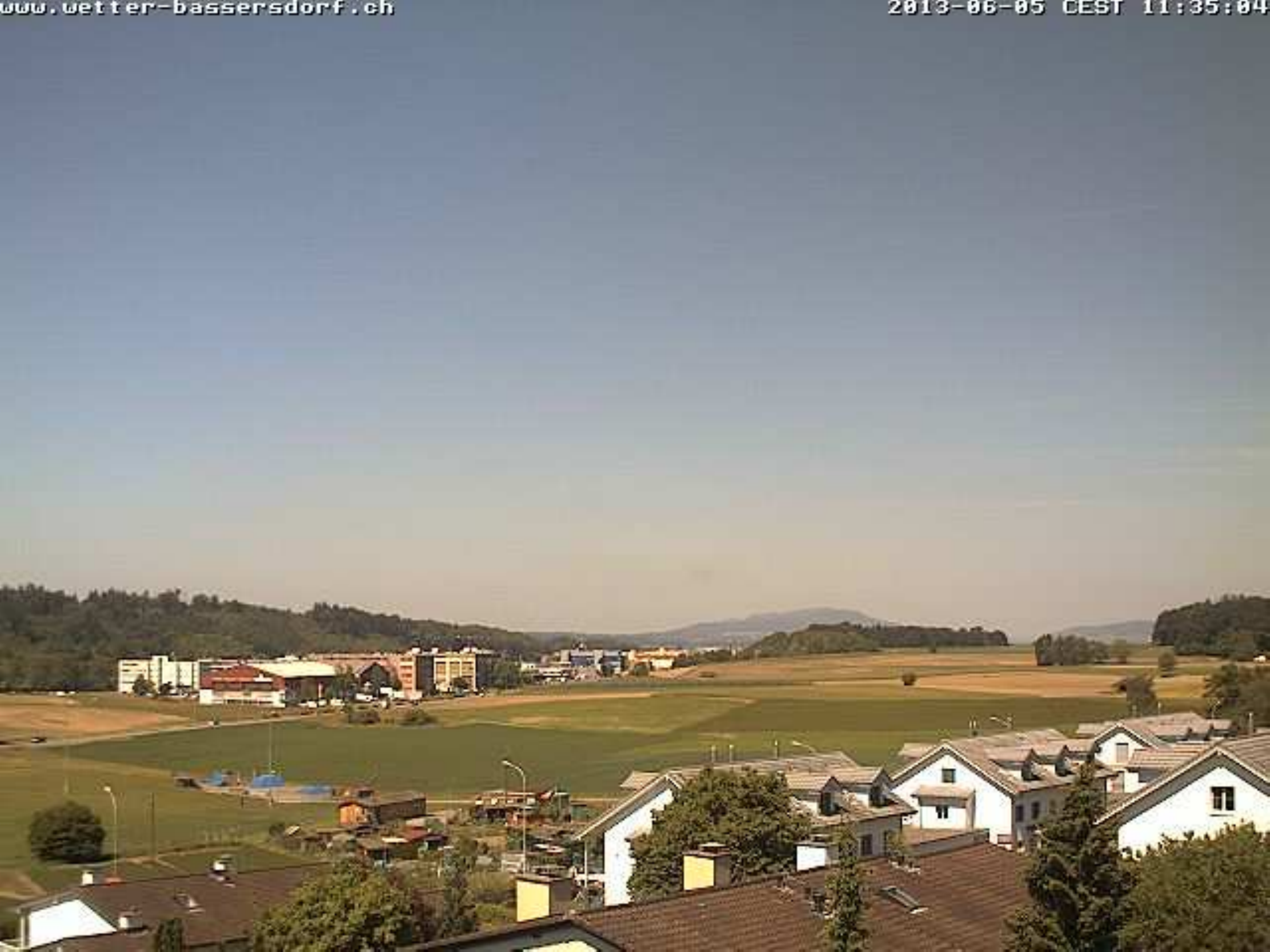} 
\includegraphics[height=2.05cm]{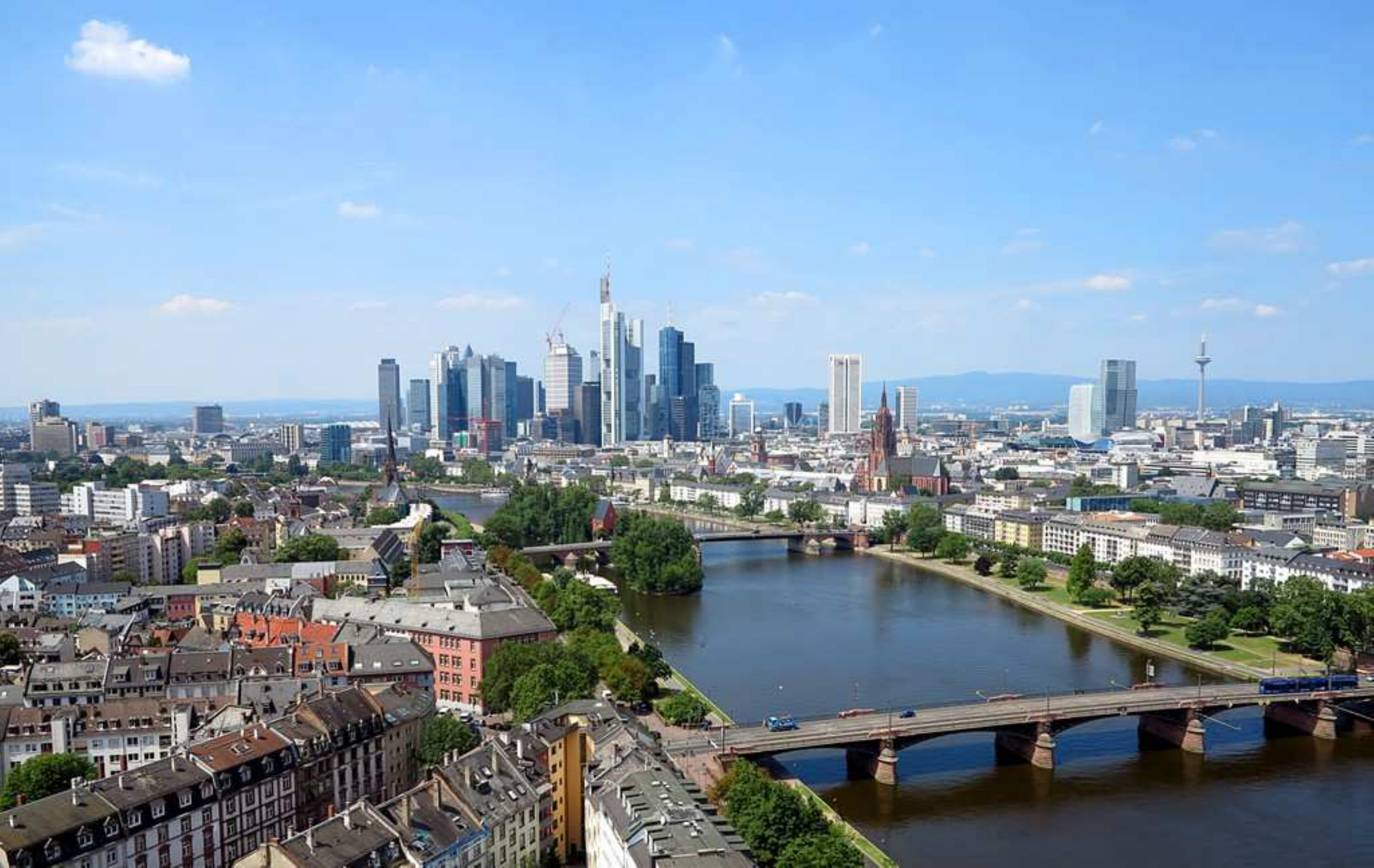} 
\includegraphics[height=2.05cm]{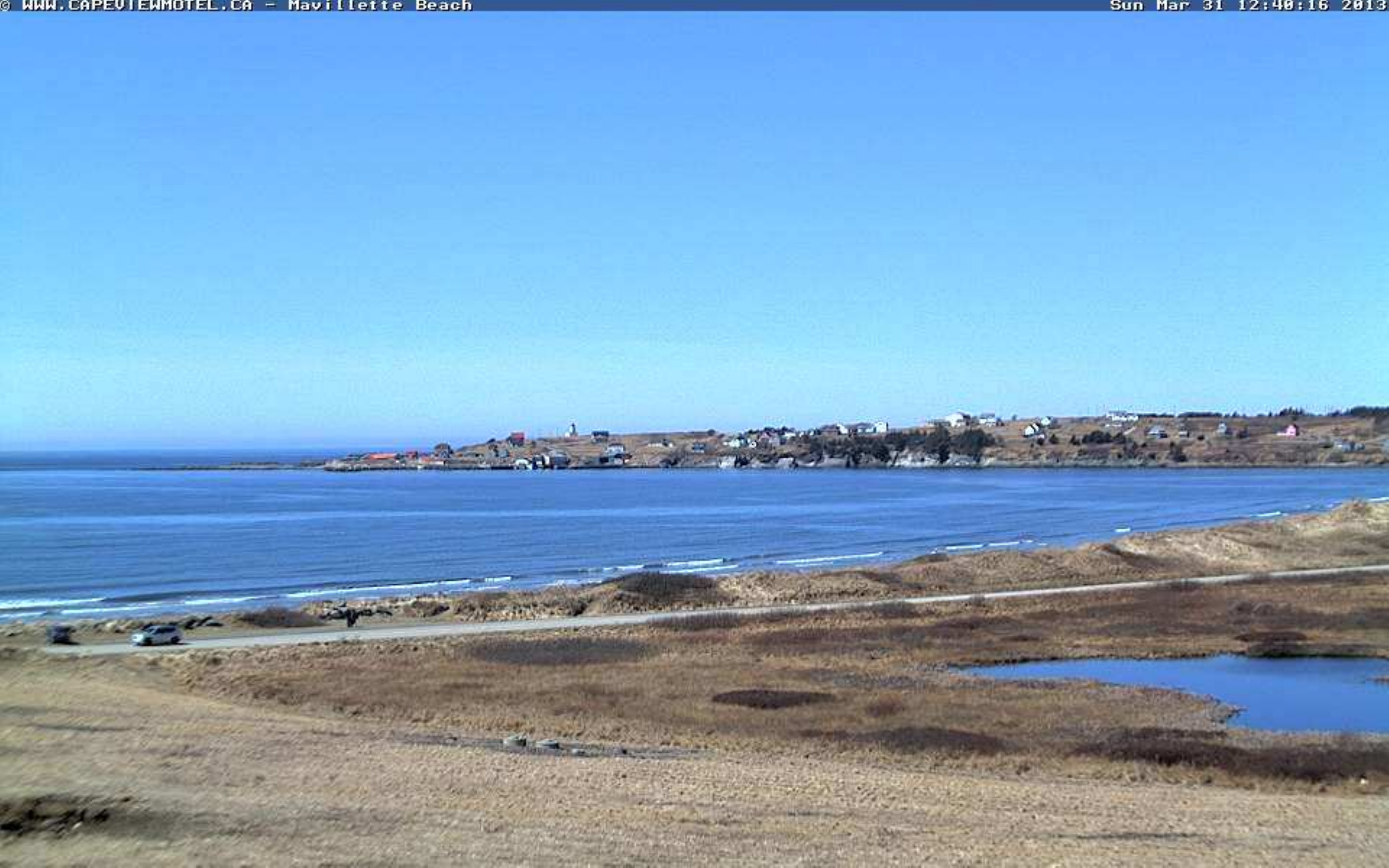} 
\caption{Sample scene images from the SkyFinder dataset \cite{mihail16}, all captured at 11 am.}
\label{fig:skyfinder}
\end{figure*}

Given an outdoor image, we aim to estimate ambient temperature based on visual information extracted from the image. We consider two application scenarios. The first one regards estimating the temperature of any given image regardless of temporal changes in the weather. In this case, only visual information extracted from the image is used. The second scenario is when several images of the same location over time are available and the goal is to \enquote{forecast} the temperature in the near future. In the first scenario, we extract visual features using a convolutional neural network (CNN), followed by a regression layer outputting the estimated temperature. In the second scenario, we also extract features using CNNs, but further consider temporal evolution by recruiting a long-short term memory (LSTM) network, which outputs the the estimated temperature of the last image in the given image sequence. 

Our contributions in this work are as follows. 
\begin{itemize}
\item We introduce a CNN-based approach to do temperature prediction for a single image and achieve promising prediction performance. 
\item We introduce a recurrent neural network (RNN) to forecast temperature of the last image in a given image sequence. To the best of our knowledge, this is the first deep model considering temporal evolution of appearance for temperature prediction. We show that our RNN-based model outperforms existing approaches. 
\item We provide a comprehensive analysis of images captured in different scenes. Interesting discussions are provided on the basis of various experimental settings and their corresponding results.  
\end{itemize}

%-------------------------------------------------------------------------
\section{Related Works}
As a pioneering work studying visual manifestations of different weather conditions, Narasimhan and Nayar \cite{nara02} firstly discussed the relationship between visual appearance and weather conditions. Since then several works have been proposed to target the weather type classification task (i.e., for weather types such as sunny, cloudy, foggy, snowy, and rainy). Roser and Moosmann \cite{roser08} analyzed the images captured by the camera mounted on vehicles, and constructed an SVM classifier to categorize images into clear, light rain, and heavy rain weather conditions. In \cite{lu17}, five weather features including sky, shadow, reflection, contrast, and haze were extracted, and a collaborative learning framework was proposed to classify images into sunny or cloudy. In \cite{chu17}, a random forest classifier was proposed to integrate various types of visual features and classify images into one of five weather conditions including sunny, cloudy, snowy, rainy, or foggy.

In addition to weather condition classification, more weather properties have also been investigated. Jacobs and his colleagues~\cite{jacobs07} initiated a project for collecting outdoor scene images captured by static webcams over a long period of time. The collected images form the Archive of Many Outdoor Scene (AMOS) dataset \cite{jacobs07}. Based on the AMOS dataset, they proposed that webcams installed across the earth can be viewed as image sensors and can enable us to understand weather patterns and variations over time \cite{jacobs09}. More specifically, they adopted principal component analysis and canonical correlation analysis to predict wind velocity and vapor pressure from a sequence of images.

Recently, Laffont et al. \cite{laffont14} estimated scene attributes like lighting, weather conditions, and seasons for images captured by webcams based on a set of regressors. Glasner et al. \cite{glasner15} studied the correlation between pixel intensity/camera motion and temperature and found a moderate correlation. With this observation, a regression model considering pixel intensity was constructed to predict temperature. Following the discussion in \cite{glasner15}, Volokitin et al. \cite{volo16} showed that, with appropriate fine tuning, deep features can be promising for temperature prediction. Zhou et al. \cite{zhou17} proposed a selective comparison learning scheme and showed that the state-of-the-art temperature prediction performance can be obtained by a CNN-based approach. 

In this work, we aim at predicting temperature from a single image, as well as forecasting the temperature of the last image in a given image sequence. Deep learning approaches will be developed to consider temporal evolution of visual appearance, and promising performance will be shown. Compared with \cite{glasner15}, \cite{volo16}, and \cite{zhou17}, we particularly advocate the importance of modeling temporal evolution with designed deep networks.

%-------------------------------------------------------------------------

\section{Data Collection and Processing}
\subsection{Datasets}
Two tasks are considered in this work. One is predicting temperature from a single image, and another is predicting the temperature of the last image in an image sequence. These two tasks are proceeded based on two datasets. For the first task, the SkyFinder dataset \cite{mihail16} consisting of a large scale webcam images annotated with sky regions is adopted. This dataset contains roughly 90,000 labeled outdoor images captured by 53 cameras in a wide range of weather and illumination conditions. By filtering out unlinked data and data without temperature information, we finally retain images captured by 44 cameras in the experiments. Currently, for each camera, only the images captured around 11am on each day are selected. Such selection decreases sunlight variations due to capturing at different time instants on a day. Finally, we are left with 35,417 images in total. We denote this image set  as \textit{Dataset 1} in what follows. Fig.~\ref{fig:skyfinder} shows five sample images from this set.

\begin{figure}[t]
\centering
 \includegraphics[width=2.7cm,height=2cm]{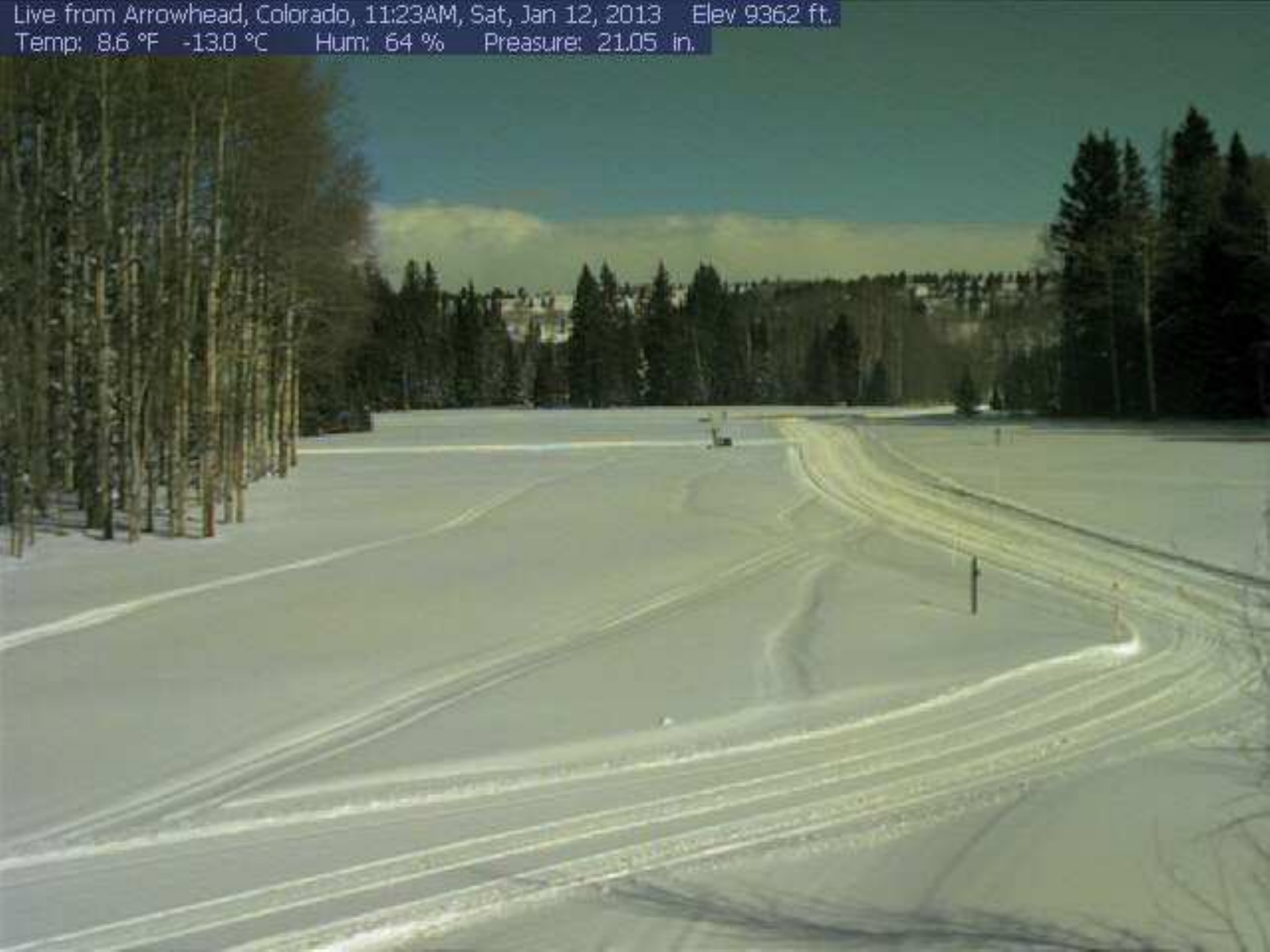} 
 \includegraphics[width=2.7cm,height=2cm]{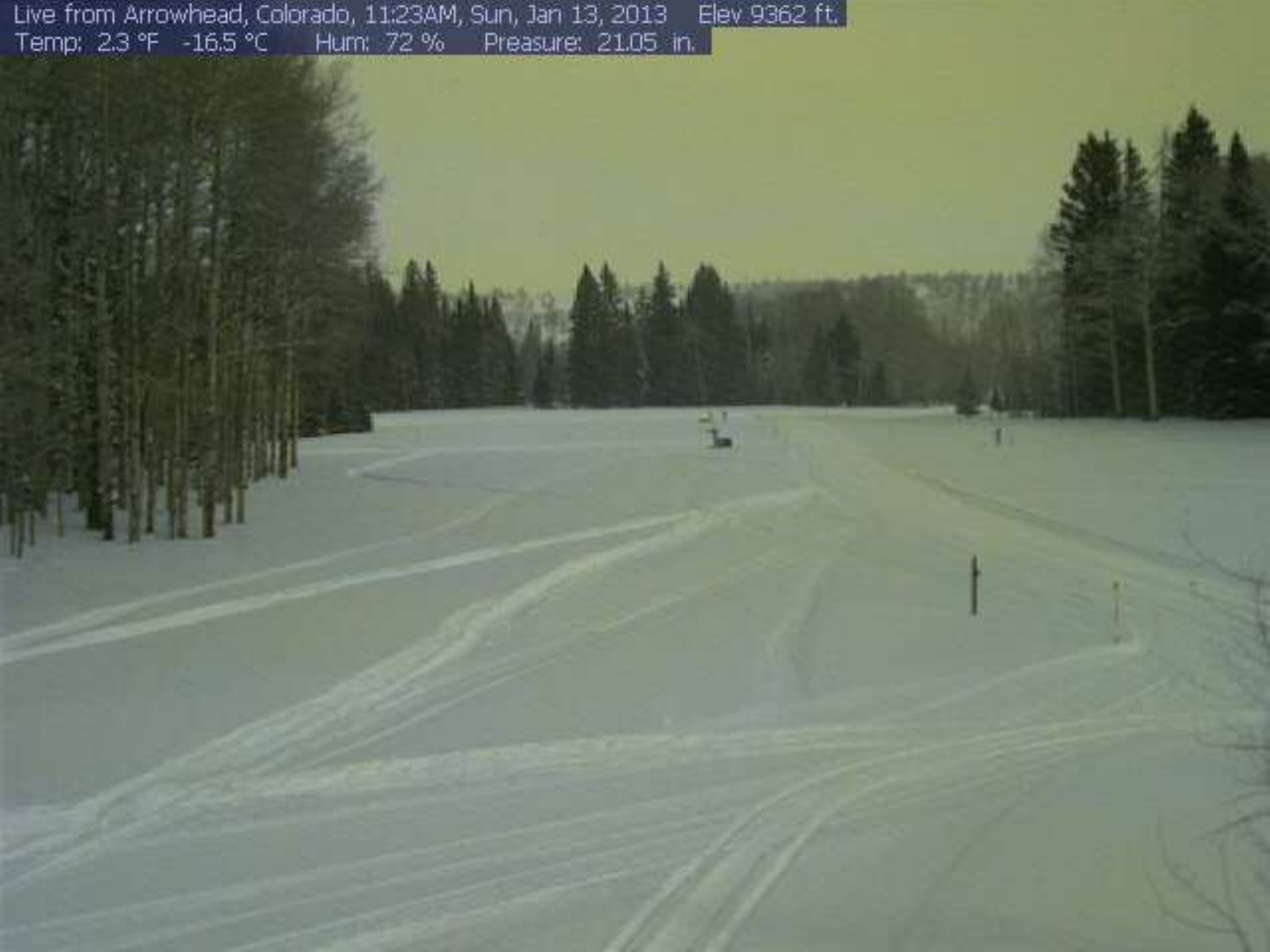} 
 \includegraphics[width=2.7cm,height=2cm]{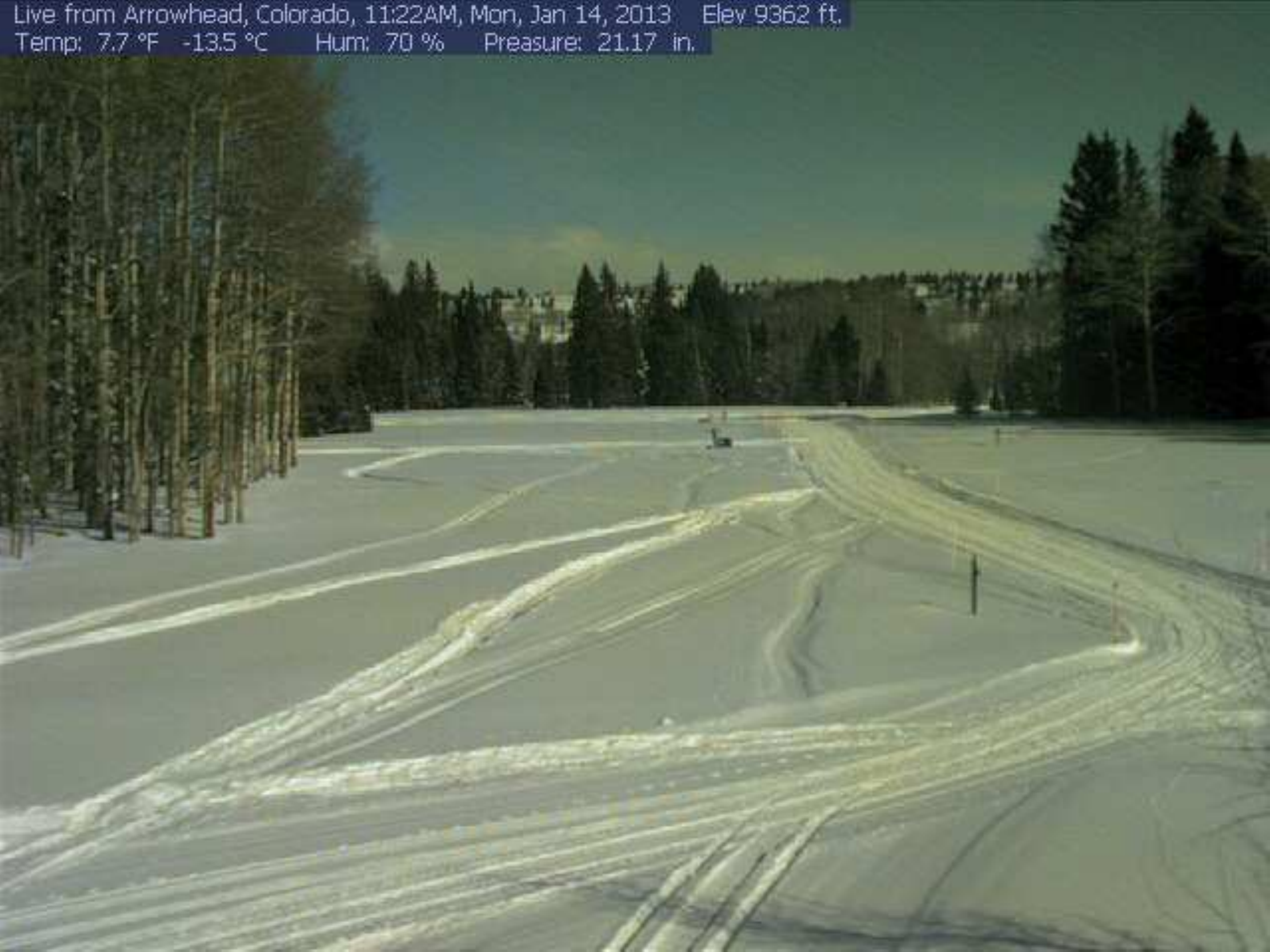} \\
 \includegraphics[width=2.7cm,height=2cm]{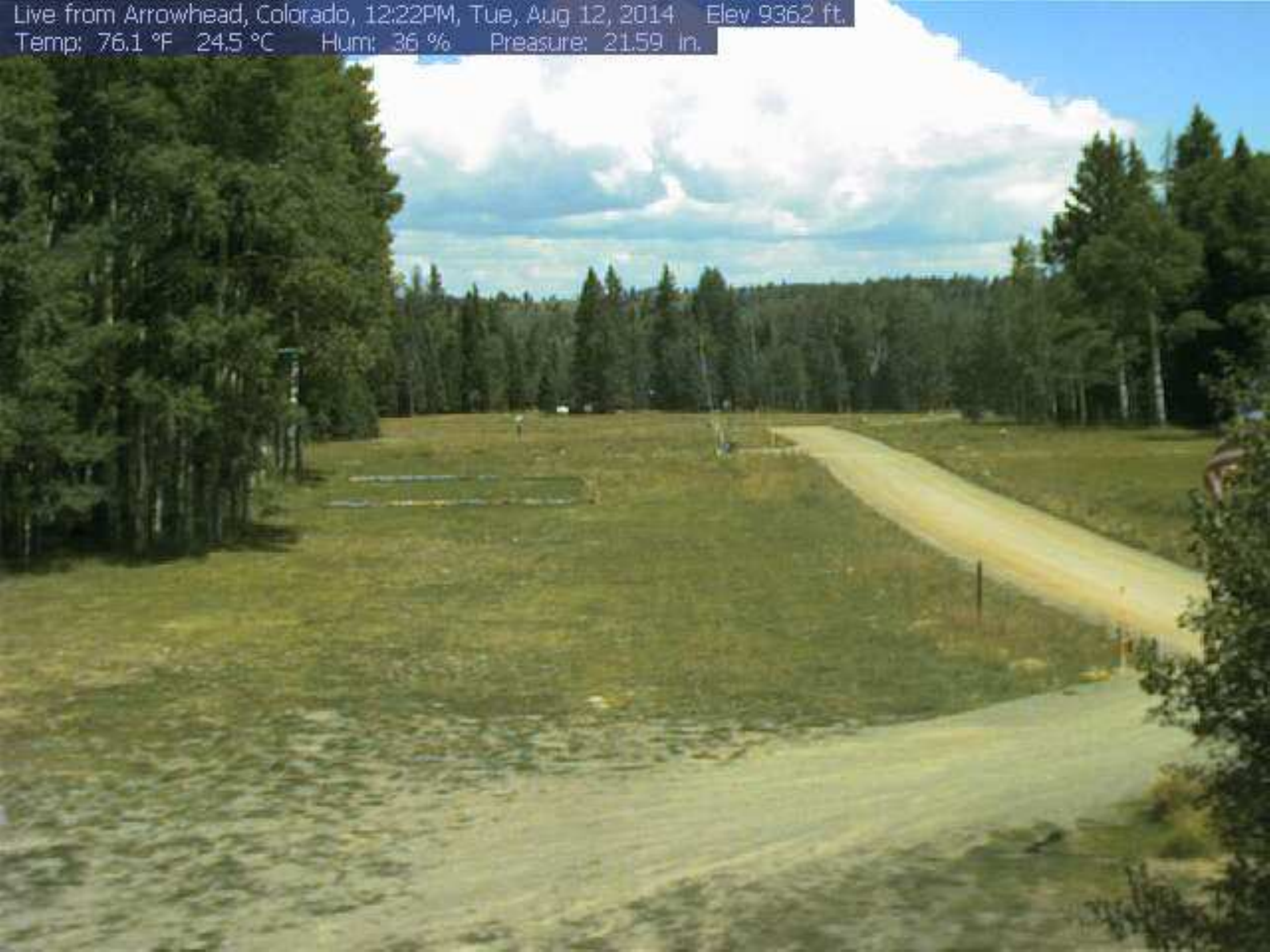} 
 \includegraphics[width=2.7cm,height=2cm]{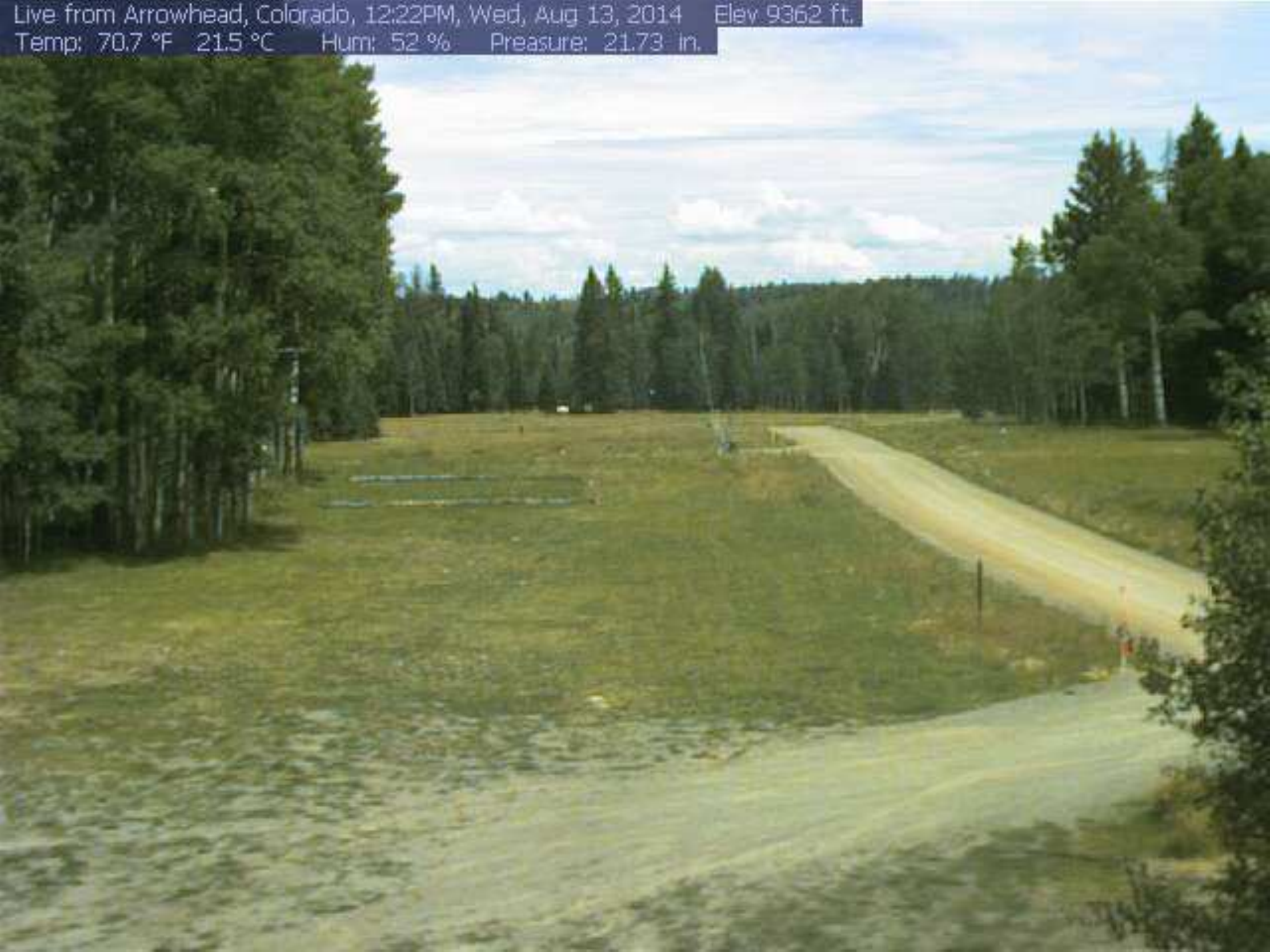}
 \includegraphics[width=2.7cm,height=2cm]{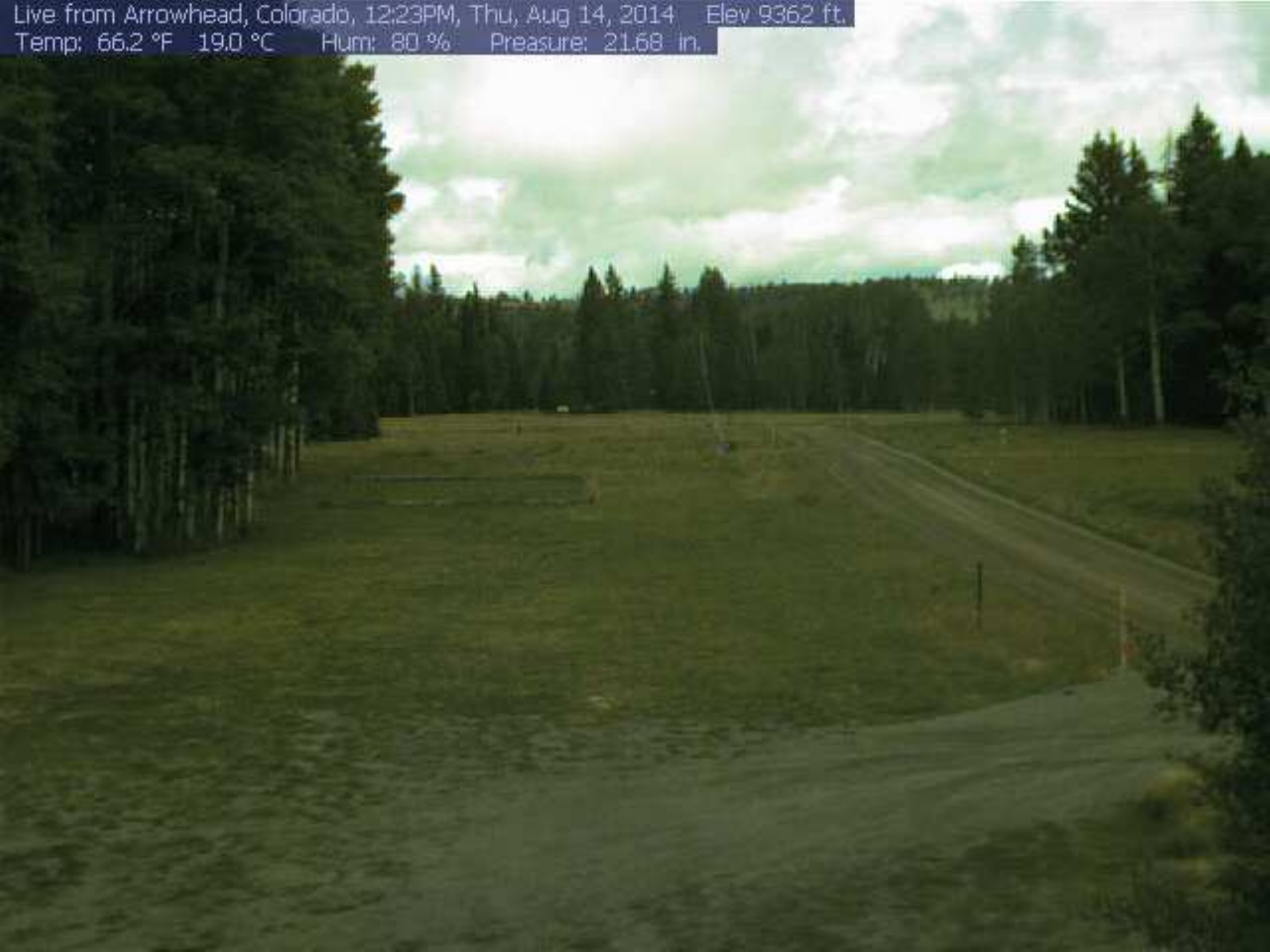} \\
 \includegraphics[width=2.7cm,height=2cm]{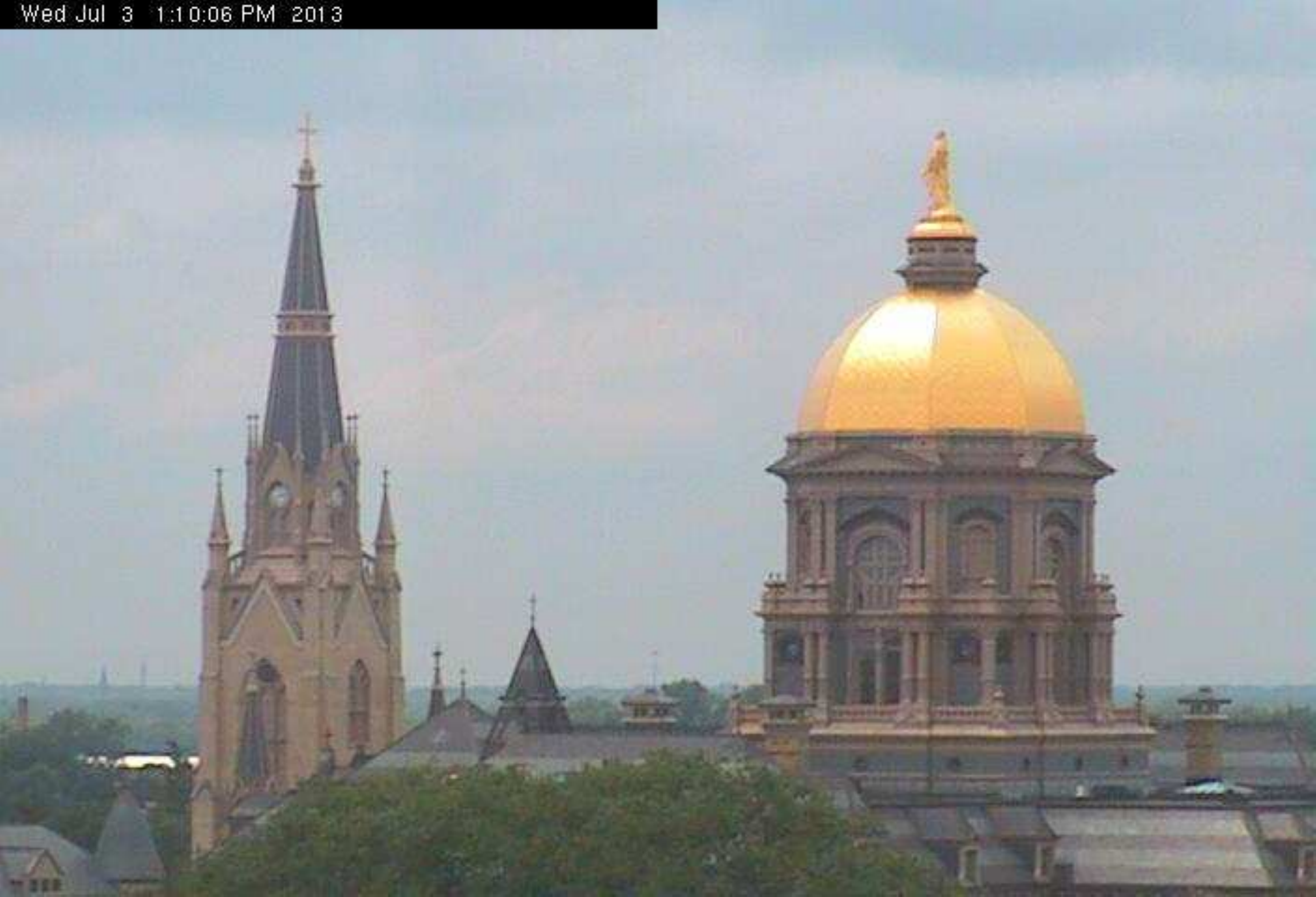} 
 \includegraphics[width=2.7cm,height=2cm]{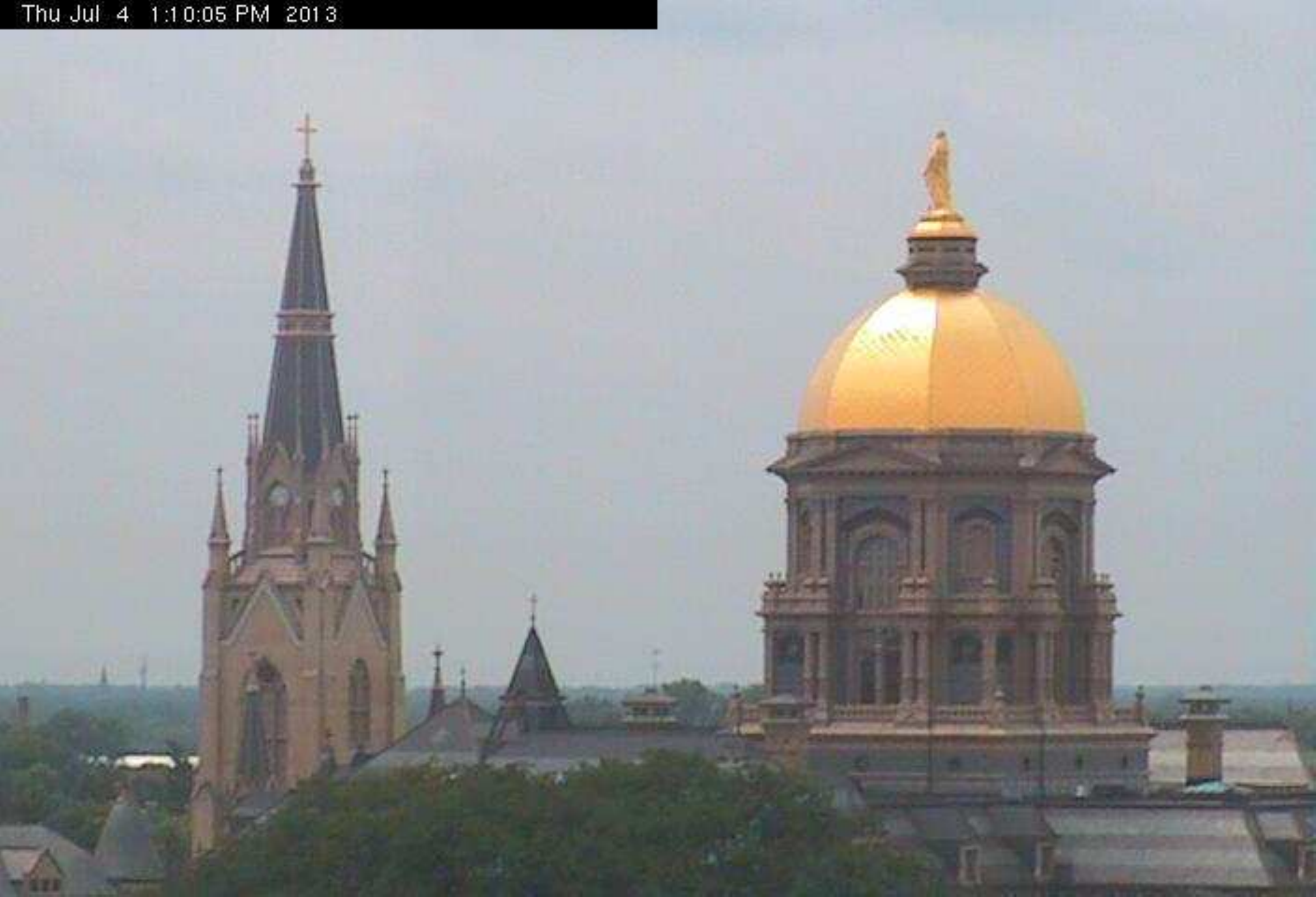}
 \includegraphics[width=2.7cm,height=2cm]{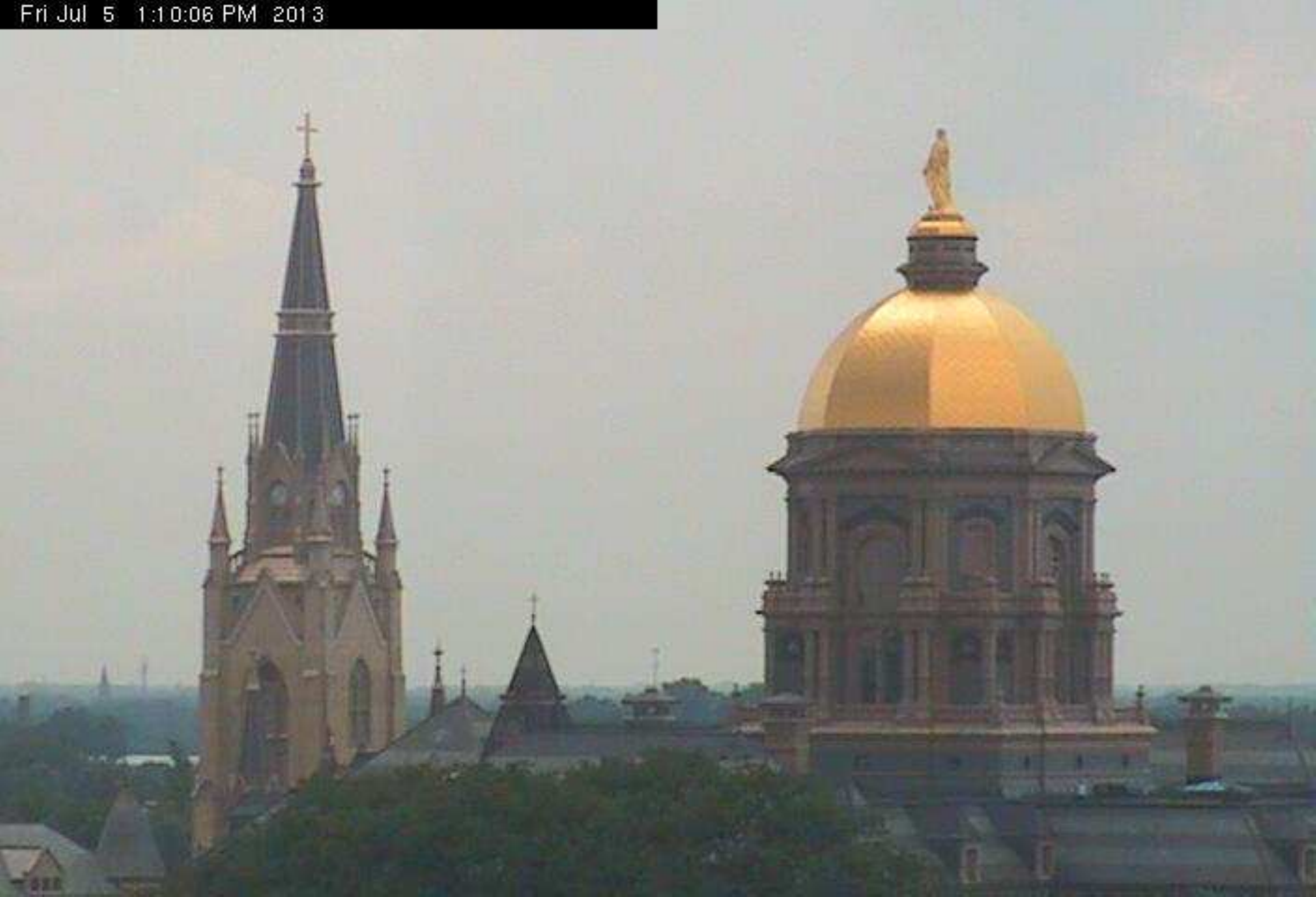} \\
 \includegraphics[width=2.7cm,height=2cm]{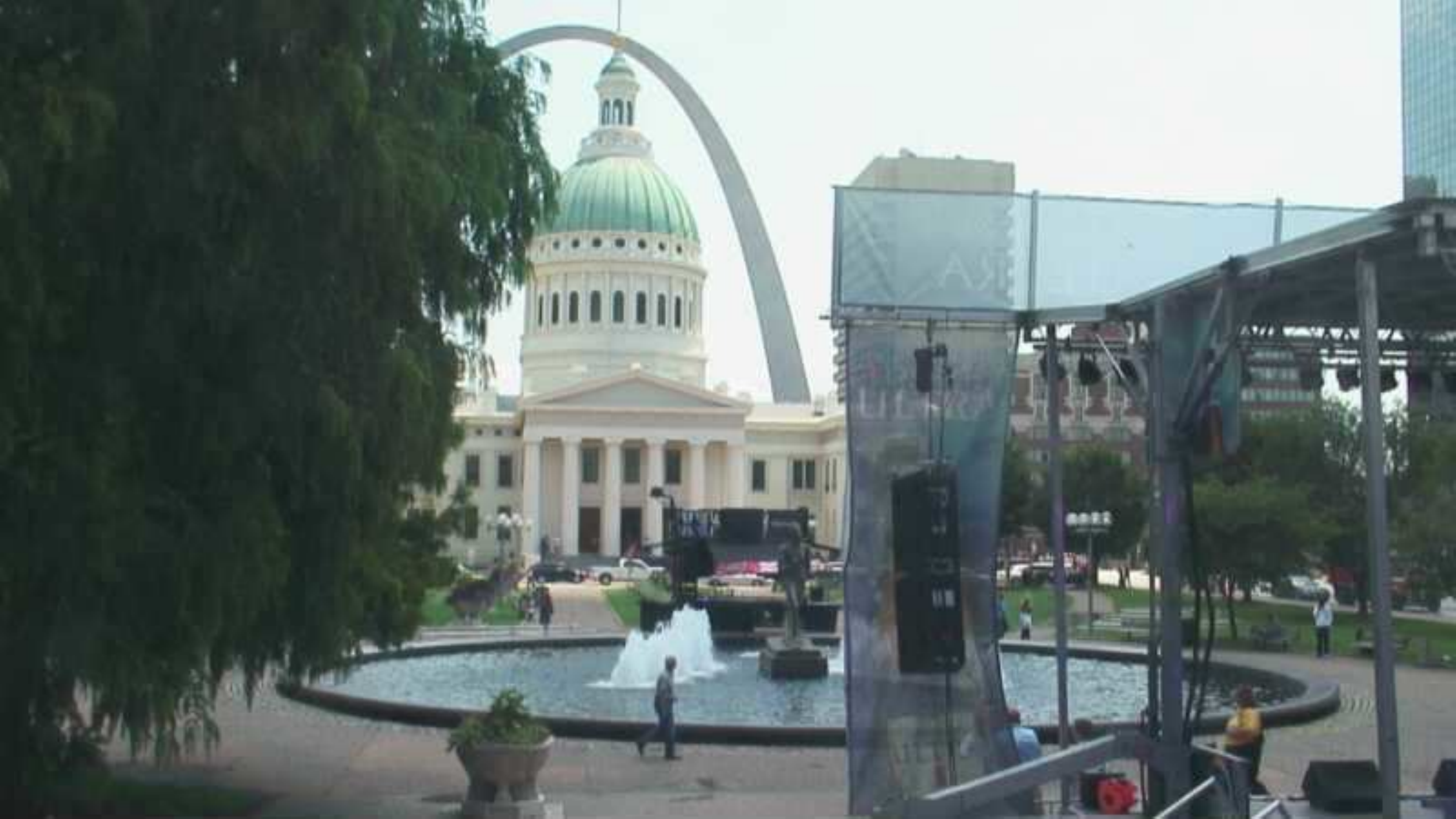} 
 \includegraphics[width=2.7cm,height=2cm]{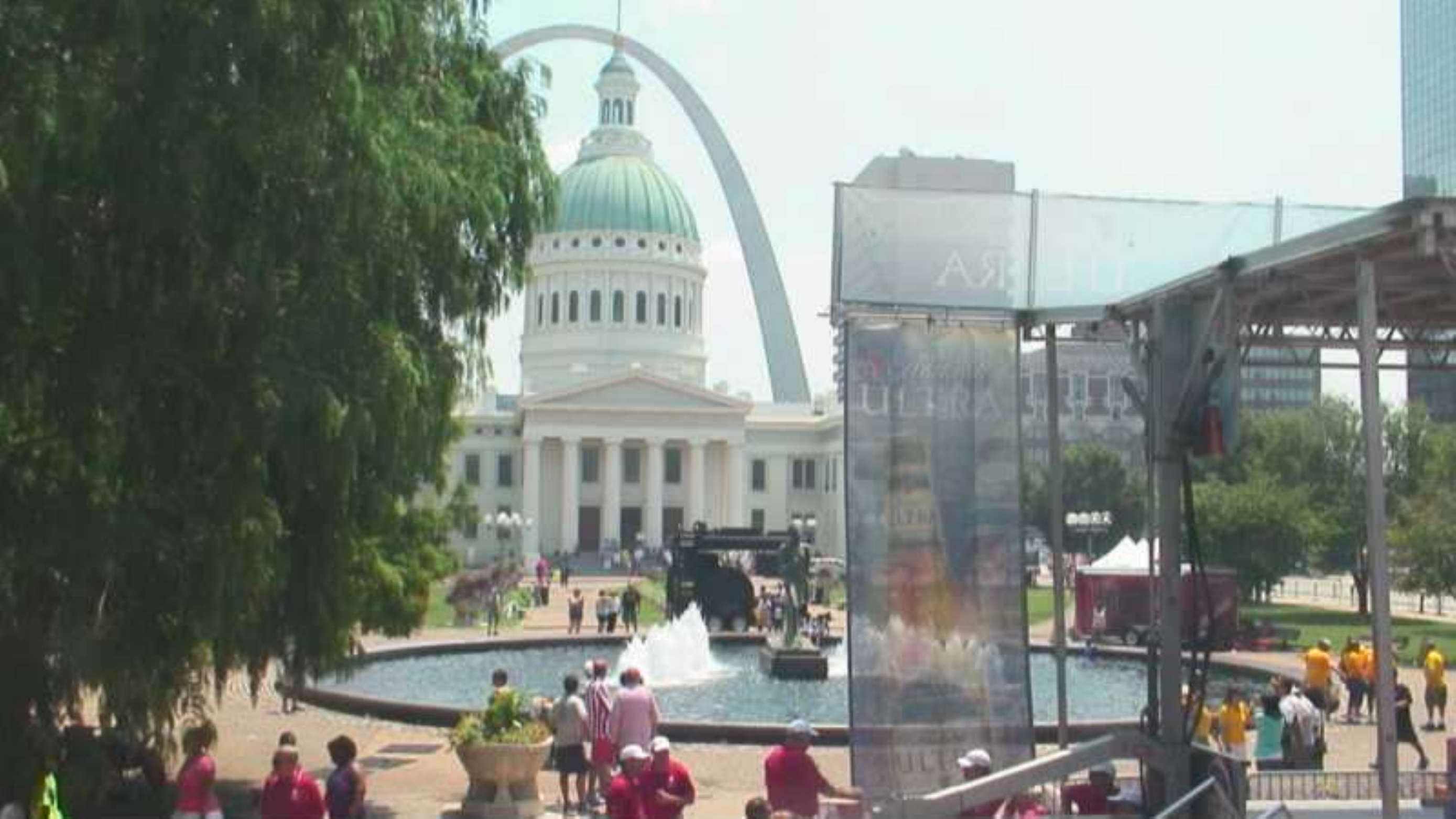}
 \includegraphics[width=2.7cm,height=2cm]{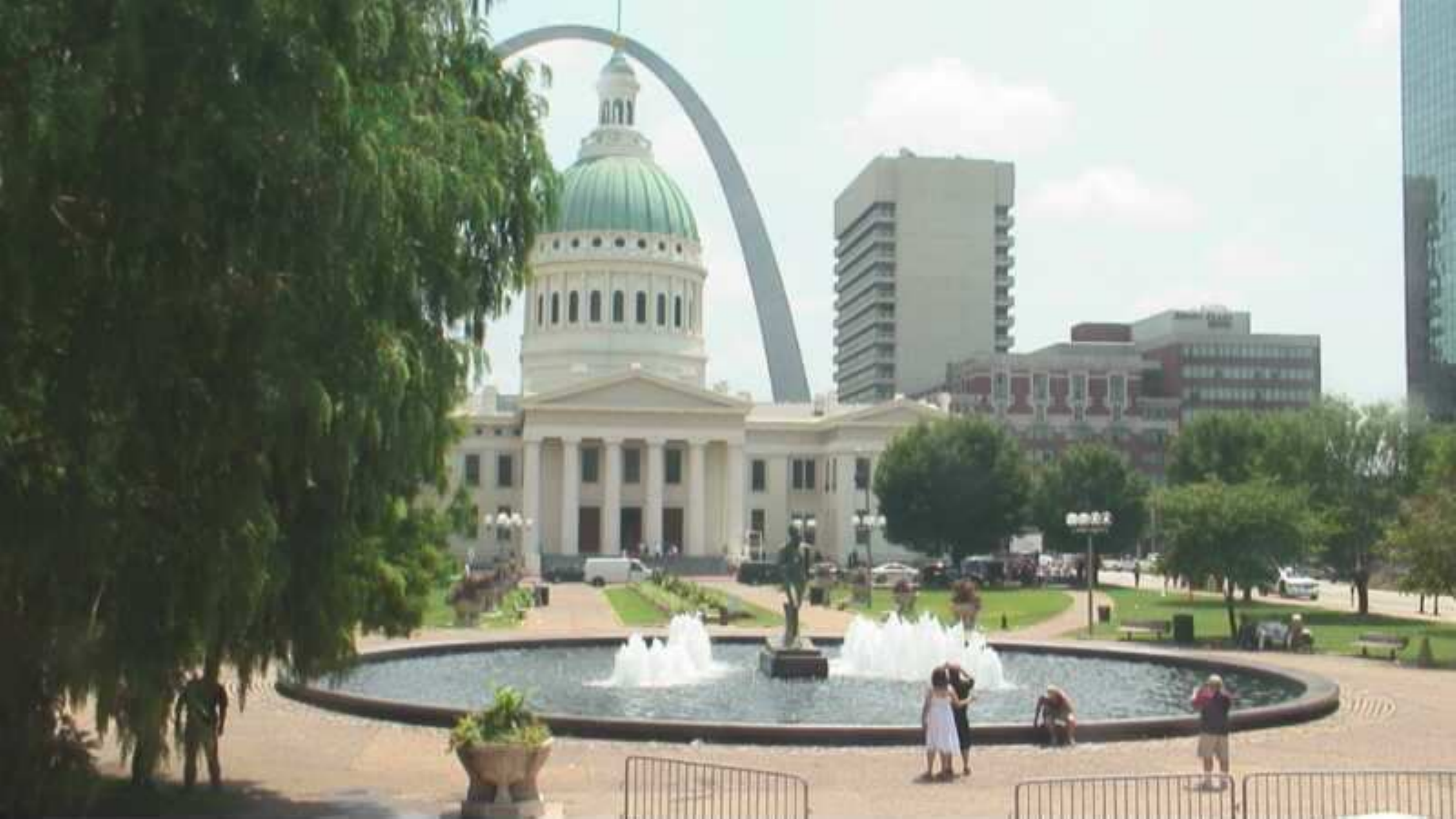}
\caption{Sample scenes from Dataset 2. First two rows correspond to the same environments in different seasons (Jan. \& Aug.).}
\label{fig:glasner}
\end{figure}

For the second task, the scene images mentioned in the Glasner dataset \cite{glasner15} are used as the seed. The Glasner dataset consists of images continuously captured by 10 cameras in 10 different environments for two consecutive years. Figure~\ref{fig:glasner} shows snapshots of 3 environments (3 for each). These cameras are in fact a small subset of the AMOS dataset \cite{jacobs07}, and from each of these ten cameras, the Glasner dataset only contains one image captured closest to 11am local time on each day. To build the proposed model, we need more data for training and testing. Therefore, according to the camera IDs mentioned in the Glasner dataset, we collect the entire set of its corresponding images from the AMOS dataset. In addition, according to the geographical information and the timestamp associated with each image, we obtain the temperature of each image from the cli-MATE website\footnote{http://mrcc.isws.illinois.edu/CLIMATE/}. Overall, we collect 53,378 images from 9 cameras\footnote{One camera's information is incorrect, and we could not successfully collect the corresponding temperature values} in total. We denote this dataset as \textit{Dataset 2} in what follows.

The first two rows of Fig.~\ref{fig:glasner} shows two different image sequences from the same scene in Dataset 2. The first row shows images captured on Jan. 12, Jan. 13, and Jan. 14, while the second row shows images captured on Aug. 14, Aug. 15, and Aug. 16. We see that visual appearance of images captured at the same scene may drastically vary due to climate changes. In addition, we clearly observe the temporal continuity of images captured on consecutive days. The third and fourth rows of Fig.~\ref{fig:glasner} show two more image sequences captured at different scenes.

\subsection{Label Encoding}
Regardless of the temperature prediction task (i.e., for a single image or an image sequence), we formulate temperature estimation as a classification problem. We divide the considered temperature range ($-20^{\circ}$C to $49^{\circ}$C) into 70 classes, represented as a 70D vector. Each dimension in this vector corresponds to a specific degree. That is, the first dimension encodes $-20^{\circ}$C, the second dimension encodes $-19^{\circ}$C, and so on. Given an image, we attempt to classify it into one of the 70 classes. We utilize two ways to encode class labels as follows. 
\begin{itemize}
\item One-hot encoding: Only the dimension representing the correct temperature is set to 1, and other dimensions are set to 0. For example, the 70D label vector $y=(0, 0, 1, 0, ..., 0)$ denotes that the corresponding image has temperature of $-18^{\circ}$C. 
\item Local distribution encoding (LDE): Motivated by \cite{antipov17}, for the image with temperature corresponding to the $i$th dimension, we set the value $t_j$ of the $j$th dimension of the label vector by a Gaussian distribution
\begin{equation}
t_j = \dfrac{1}{\sigma \sqrt{2\pi}} \exp(\frac{(j-i)^2}{2 \sigma^2}). 
\end{equation} 
\end{itemize} 

We will compare two encoding schemes in the evaluation section. 

\begin{table*}
\centering
\caption{Detailed configurations of the convolutional neural network. }
\renewcommand{\arraystretch}{1.4}
\renewcommand{\tabcolsep}{1mm}
\begin{tabular}{ c|c |c |c |c |c |c }
\hline
Conv2D(32, 3, 3) & \pbox{3cm}{Conv2D(32, 3, 3) \\ MaxPooling2D(2, 2) \\ Dropout(0.25)} & Conv2D(64, 3, 3) & \pbox{3cm}{Conv2D(64, 3, 3) \\ MaxPooling2D(2, 2) \\ Dropout(0.25)} & Flatten & \pbox{3cm}{Dense(512) \\ Dropout(0.5)} & \pbox{3cm}{Dense(70) \\ Softmax}\\
\hline
\end{tabular}
\label{tb:model}
\end{table*} 

\section{Proposed Methods}
This section presents details of the proposed models for temperature prediction. We are interested in predicting the temperature for a given still image as well as predicting the temperature of a location from temperature data of the same location at the same time on previous days (i.e., next day prediction).

%-------------------------------------------------------------------------
\subsection{Temperature Estimation from a Single Image}
\label{sec:single}
To estimate ambient temperature of a given outdoor image, we construct a convolutional neural network (CNN) trained from scratch based on the SkyFinder dataset. Table~\ref{tb:model} shows detailed configurations of our CNN architecture. There are totally 4 convolutional layers, followed by 4 fully-connected layers. The model's output is a 70D vector indicating the probabilities of different temperatures (i.e., classes). To train the model, the activation function of each layer is ReLU, the loss function is cross entropy, the optimization algorithm is Adam, and learning rate is 0.001. Different batch sizes are tested to achieve the best performance for different experimental settings. 80\% of the SkyFinder dataset is used for training, and the remaining 20\% is used for testing. We do not use a validation set and train the network for 90 epochs.  

%-------------------------------------------------------------------------

\subsection{Temperature Estimation from a Sequence of Images}
\label{sec:rnn}
Given an image sequence $I_1, I_2, ..., I_n$ captured at the same scene, and assuming that the corresponding temperature values of the first $n-1$ images, i.e., $\boldsymbol{t}_1, \boldsymbol{t}_2, ..., \boldsymbol{t}_{n-1}$ are available, we would like to predict the temperature $\boldsymbol{t}_n$ of the image $I_n$. In addition to the visual information of $I_n$, temporal evolution of the visual appearance of $I_1$ to $I_{n-1}$ may provide information regarding future temperatures. Therefore, we construct a long-short term memory network (LSTM) \cite{hoch97} to successively propagate visual information over time to predict temperature.

\begin{figure}
\centering
\includegraphics[width=8cm]{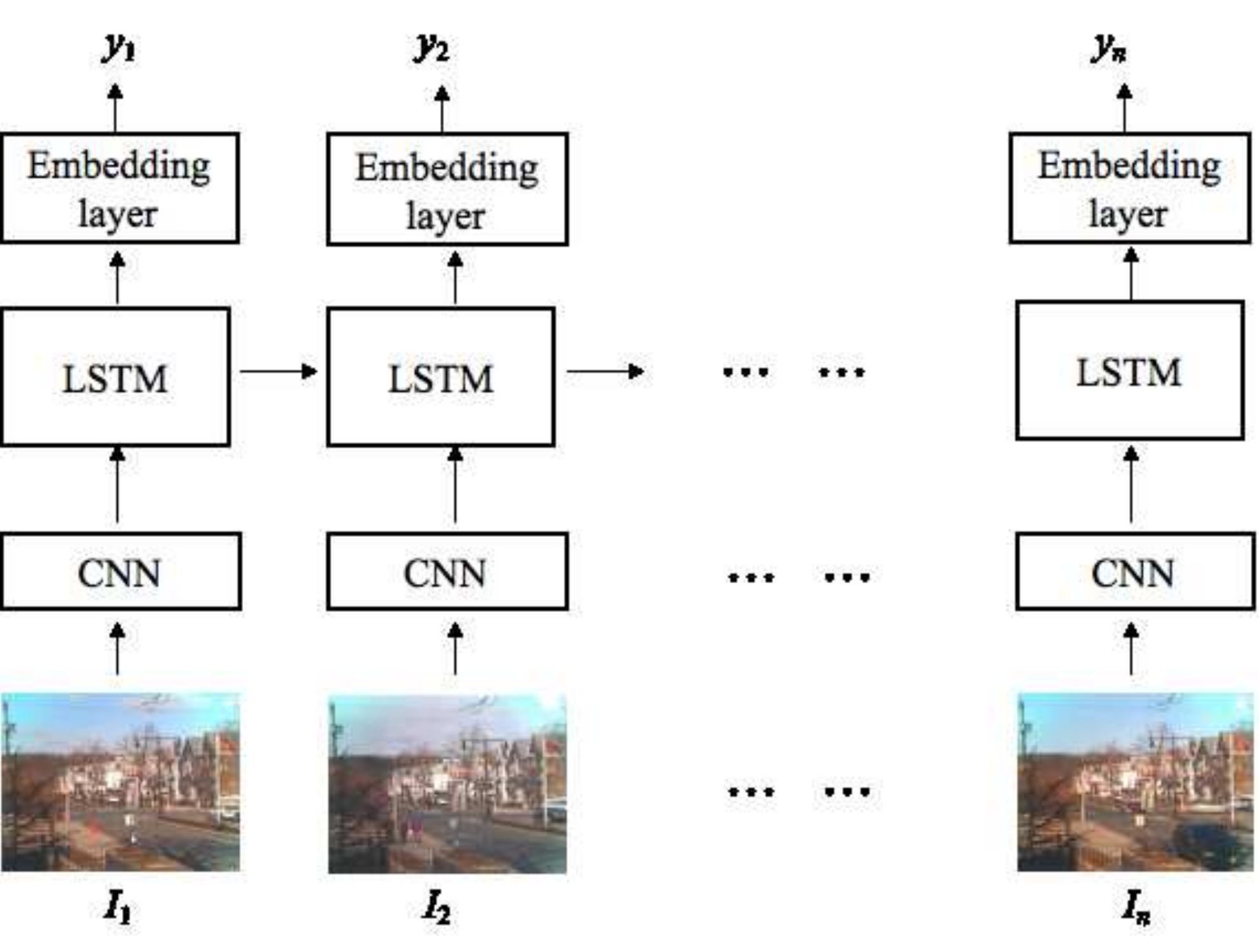} 
\caption{Structure of the proposed LSTM network for temperature prediction. }
\label{fig:RNN}
\end{figure}

Fig.~\ref{fig:RNN} shows the model structure. Each image in the sequence is first fed to a CNN to extract visual features. This CNN has the same structure as mentioned in in Sec.~\ref{sec:single}, without the last softmax layer. The extracted feature vector of the image $I_i$ is then fed to an LSTM layer, which not only processes the current input, but also considers the information propagated from the intermediate result for the image $I_{i-1}$. Similarly, the intermediate result for the image $I_i$ will be sent to the LSTM layer for processing the image $I_{i+1}$. The output of the LSTM layer would be input to an embedding layer that transforms the input vector into a 70D vector $\boldsymbol{y}_i$ indicating the probabilities of different temperatures. 

To train the RNN, the loss function is defined as the mean square error between the ground truth and the predicted vector. That is, 
\begin{equation}
L = \sum_{i=1}^n \lVert \boldsymbol{y}_i - \boldsymbol{t}_i \rVert^2. 
\end{equation}

Fig.~\ref{fig:RNN} shows that information is only propagated from $I_1$ to $I_n$. In addition to this forward propagation scheme, we also try to construct a bi-directional LSTM model to consider both forward propagation and backward propagation. Detailed performance comparison will be shown later. 

%-------------------------------------------------------------------------
\section{Empirical Evaluation}
\subsection{Implementation Details} 
\label{sec:implement}
To estimate ambient temperature of a single image, the CNN model mentioned in Sec.~\ref{sec:single} is constructed based on 80\% of the images in Dataset 1, while the remaining 20\% of the images are used for testing. We repeat this process five times with the five-fold cross validation scheme, and report the average estimation error. In this task, the standard deviation $\sigma$ mentioned in the LDE scheme is empirically set as 3.5. 

More details for predicting the temperature of the last image of an image sequence are provided as follows. For a camera, let ${I_{08}^{(t)}, I_{09}^{(t)}, ..., I_{17}^{(t)}}$ denote the images with timestamps closest to 8am, 9am, ..., 17pm, on day $t$. To train the RNN mentioned in Sec.~\ref{sec:rnn}, we select a sequence consisting of $n$ images that were captured around the same time on $n$ consecutive days, like $(I_{08}^{(t)}, I_{08}^{(t+1)}, ..., I_{08}^{(t+n-1)})$ and $(I_{15}^{(t)}, I_{15}^{(t+1)}, ..., I_{15}^{(t+n-1)})$. We always keep images captured closest to 11am as the testing data, and exclude them in the training data set. The training image sequences collected from all cameras are put together to be the generic training dataset, and the (generic) RNN is trained based on it. In this task, the standard deviation $\sigma$ mentioned in the LDE scheme is empirically set as 4.

Given a test image sequence $(I_{11}^{(t)}, I_{11}^{(t+1)}, ..., I_{11}^{(t+n-1)})$, assuming that we have known the temperature values of $(I_{11}^{(t)}, I_{11}^{(t+1)}, ..., I_{11}^{(t+n-2)})$, we would like to predict the temperature values of $I_{11}^{(t+n-1)}$. In our experiments, we found that the best performance can be obtained when $n$ is set as three. That is, given the images captured on day $t$ and day $t+1$, we predict the temperature of the image captured on day $t+2$.

\begin{table}[b]
\centering
\caption{Root mean square errors when different percentages of data in Dataset 1 used during training. } 
\begin{tabular}{c |c |c |c |c }
\hline
 & 20\% & 40\% & 60\% & 80\%  \\
\hline
Avg. RMSE & 5.34 & 4.86 & 4.55 & \textbf{4.28} \\
\hline
\end{tabular}
\label{tb:volume}
\end{table}

\begin{table*}[t]
\centering
\renewcommand{\tabcolsep}{1.9mm}
\caption{Temperature estimation errors, in terms of RMSE, for each of the 9 scenes in the Glasner dataset. }
\begin{tabular}{c |c |c |c |c |c |c |c |c |c |c ||c}
\hline
 {\bf Method } $\downarrow$  $\backslash$ {\bf Scene} $\rightarrow$ & (a) & (b) & (c) & (d) & (e) & (f) & (g) & (h) & (i) & (j) & Avg. \\
\hline
\hline
Local Regression \cite{glasner15} & 6.85 & 7.24 & 6.03 & 4.53 & 5.77 & 3.19 & 7.63 & 5.09 & 3.68 & 7.77 & 5.78\\
\hline
LR Temporal Win. \cite{glasner15} & 7.52 & 6.86 & 5.82 & 4.23 & 5.39 & 2.77 & 7.35 & 5.22 & 3.67 & 7.85 & 5.67\\
\hline
Global Ridge Reg. \cite{glasner15} & 18.16 & 5.74 & 35.02 & 11.37 & 43.51 & 3.84 & 5.54 & 13.86 & 3.41 & 8.91 & 14.94\\
\hline
CNN \cite{glasner15} & 8.55 & 5.59 & 5.96 & 6.17 & 7.30 & 2.90 & 8.48 & 4.88 & 2.93 & 7.12 & 5.99\\
\hline
Fine-tuned VGG-16 \cite{volo16} & 7.79 & 4.87 & 5.03 & 4.45 & 4.22 & 3.14 & 6.61 & 4.72 & 2.70 & 6.01 & 4.96\\
\hline
Selective Comparison Learning \cite{zhou17} & 6.26 & 4.57 & 4.92 & \textbf{3.80} & \textbf{3.98} & \textbf{2.53} & \textbf{5.20} & 3.95 & \textbf{2.48} & \textbf{5.81} & \textbf{4.35}\\
\hline
Our CNN (LDE) & \textbf{4.38} & \textbf{3.79} & \textbf{4.34} & 4.60 & 4.47 & 5.83 & 5.36 & \textbf{3.73} & 2.81 & 6.47 & 4.58\\
\hline
Our CNN (one hot) & 6.28 & 5.48 & 5.78 & 6.69 & 6.36 & 7.69 & 6.78 & 6.02 & 3.25 & 7.54 & 6.19\\
\hline
\end{tabular}
\label{tb:cnnperf_com}
\end{table*} 

\subsection{Performance of Temperature Estimation for Single Images}
Similar to \cite{glasner15}, \cite{volo16}, and \cite{zhou17}, we evaluate the prediction performance based on the root mean squared error, RMSE, between the estimated temperature and the ground truth. The dataset used in  \cite{glasner15}, \cite{volo16}, and \cite{zhou17} is relatively small, i.e., images captured by ten cameras, and from each day only one image is selected. This may impede the development of a robust model. For example, in \cite{glasner15}, the authors report that the CNN-based approach does not work very well, but in \cite{volo16}, the authors show that more promising performance can be obtained by appropriately fine-tuning a pre-trained CNN model, i.e., VGG-16 \cite{simon14}. 

\textbf{Volume of training data.} In this evaluation, we first focus on studying how the volume of training data influences temperature estimation performance, based on a larger-scale dataset, i.e., Dataset 1. We train 4 CNN models on 20\%, 40\%, 60\%, and 80\% of the data in Dataset 1, respectively, and use the same remaining 20\% of data for testing. As shown in Table~\ref{tb:volume}, unsurprisingly more training data yield better performance. 

\textbf{Performance comparison.} To fairly compare our CNN model with existing works, we train the proposed CNN model on the Glasner dataset. Note that volume of the Glasner dataset is smaller than Dataset 1. 
Table~\ref{tb:cnnperf_com} shows comprehensive performance comparison, where ten cameras in the Glasner dataset are referred to as scenes (a)-(j). We find that results of \cite{volo16} are superior to \cite{glasner15}, which demonstrates fine-tuning a deep model yields performance gain. Comparing our CNN (with the LDE scheme) with \cite{volo16}, our  CNN is much simpler than VGG-16 and achieves even better results. The performance gain might be due to the LDE scheme, which may provide more flexibility to model temperature fluctuation. To verify this, we show performance obtained by our CNN model with the one-hot encoding scheme in the last row of Table~\ref{tb:cnnperf_com}. By comparing the last two rows, we clearly see the benefits brought by the LDE scheme. Performance of the work in \cite{zhou17} remains the best. A two-stream network was proposed to take an image pair captured at the same place and on the same day as input. Each network stream extracts features from one image, and estimates the temperature of each image. To train this two-stream network, in addition to the softmax losses given by each stream, a ranking loss representing the temperature difference between two images is jointly considered. Our CNN model with LDE fares on par with \cite{zhou17} but is much simpler. Further, the idea of ranking loss can be integrated in our model to make performance gain in the future. 

\subsection{Performance of Temperature Estimation for Image Sequences}
\textbf{Length of the image sequence.} We first evaluate estimation performance when different lengths of image sequences are used for training and testing, and show the average RMSEs in Table~\ref{tb:length}. As can be seen, the best estimation performance can be obtained when $n$ is set as 3. That is, we estimate the temperature of the day $t+2$ based on day $t+1$ and day $t$. This result is not surprising because this setting appropriately considers information of previous days, and prevents blunt updates when too many days are considered. Therefore, we set $n=3$ in the subsequent experiments.  

\begin{table}
\centering
\caption{Root mean square errors when different lengths of image sequences are used. } 
\begin{tabular}{c |c |c |c |c |c |c }
\hline
 & n=2 & n=3 & n=4 & n=5 & n = 6 & n=7 \\
\hline
Avg. RMSE & 2.97 & \textbf{2.82} & 2.95 & 2.87 & 2.86 & 2.98 \\
\hline
\end{tabular}
\label{tb:length}
\end{table}

\begin{table*}[htbp!]
\centering
\caption{Temperature estimation errors (in terms of RMSE) for each scene in the Glasner dataset.}
\begin{tabular}{c |c |c |c |c |c |c |c |c |c |c ||c}
\hline
 {\bf Method } $\downarrow$  $\backslash$ {\bf Scene} $\rightarrow$  & (a) & (b) & (c) & (d) & (e) & (f) & (g) & (h) & (i) & (j) & Avg. \\
\hline
\hline
Freq. Decom. \cite{glasner15} & \textbf{5.05} & 6.15 & 4.61 & 4.10 & 5.27 & 2.51 & 4.51 & 4.50 & 2.81 & 5.16 & 4.47\\
\hline
Our single LSTM & 6.01 & \textbf{2.32} & 2.53 & -- & 2.26 & \textbf{2.46} & 2.93 & 2.04 & 1.62 & 3.02 & \textbf{2.80}\\
\hline
Our bidirectional LSTM & 7.09 & 2.54 & \textbf{2.49} & -- & \textbf{1.93} & 2.61 & \textbf{2.36} & \textbf{1.89} & \textbf{1.57} & \textbf{2.93} & \textbf{2.82}\\
\hline
\hline
Handcrafted + our bi-LSTM & 10.84 & 6.60 & 11.79 & -- & 8.91 & 11.71 & 7.93 & 9.47 & 3.50 & 9.58 & 8.93\\
\hline
\end{tabular}
\label{tb:rnnperf_com}
\end{table*} 

\textbf{Performance comparison.} Table~\ref{tb:rnnperf_com} shows performance of the proposed single LSTM and bidirectional LSTM, and compares performance with the state of the art methods. Previously only the frequency decomposition method proposed in \cite{glasner15} takes temporal evolution into account. Note that the data used in \cite{glasner15} are from the Glasner dataset. Our LSTM models are trained and tested on Dataset 2 that is also from the Glasner dataset but includes much more images to capture temporal evolution. We discarded scene (d) as its data have errors in the geographical information and we could not access corresponding temperature information. As can be seen from Table~\ref{tb:rnnperf_com}, excluding camera (d), the RNN-based approach achieves better performance over eight of the nine scenes\footnote{The average RMSE of the frequency decomposition method is calculated based on ten scenes, while that of our LSTM-based methods is calculated based on nine scenes.}. The proposed RNN-based approach outperforms the state of the art. The frequency decomposition method attempts to model the temporal evolution of the entire year, while we consider local temporal evolution and achieve much better performance. 

The structure shown in Fig.~\ref{fig:RNN} shows that the LSTM model works based on features extracted by the CNN designed like Table~\ref{tb:model}. To demonstrate the influence of the CNN, we replace the CNN features by a handcrafted feature, say Gabor wavelet texture feature \cite{man96}, and train a bidirectional LSTM based on it. The performance obtained based on the handcrafted features is shown in the last row of Table~\ref{tb:rnnperf_com}. The average RMSE is 8.93, which is much higher than 2.80 and evidently demonstrates the effectiveness of CNN features. 

Overall, Table~\ref{tb:rnnperf_com} shows that temporal evolution of visual appearance provides rich information in estimating ambient temperature, and yields substantially better estimation performance. To our knowledge, this result is the first work considering temporal evolution by a deep neural network to do ambient temperature estimation. 

\begin{figure}
\centering
 \includegraphics[width=4cm]{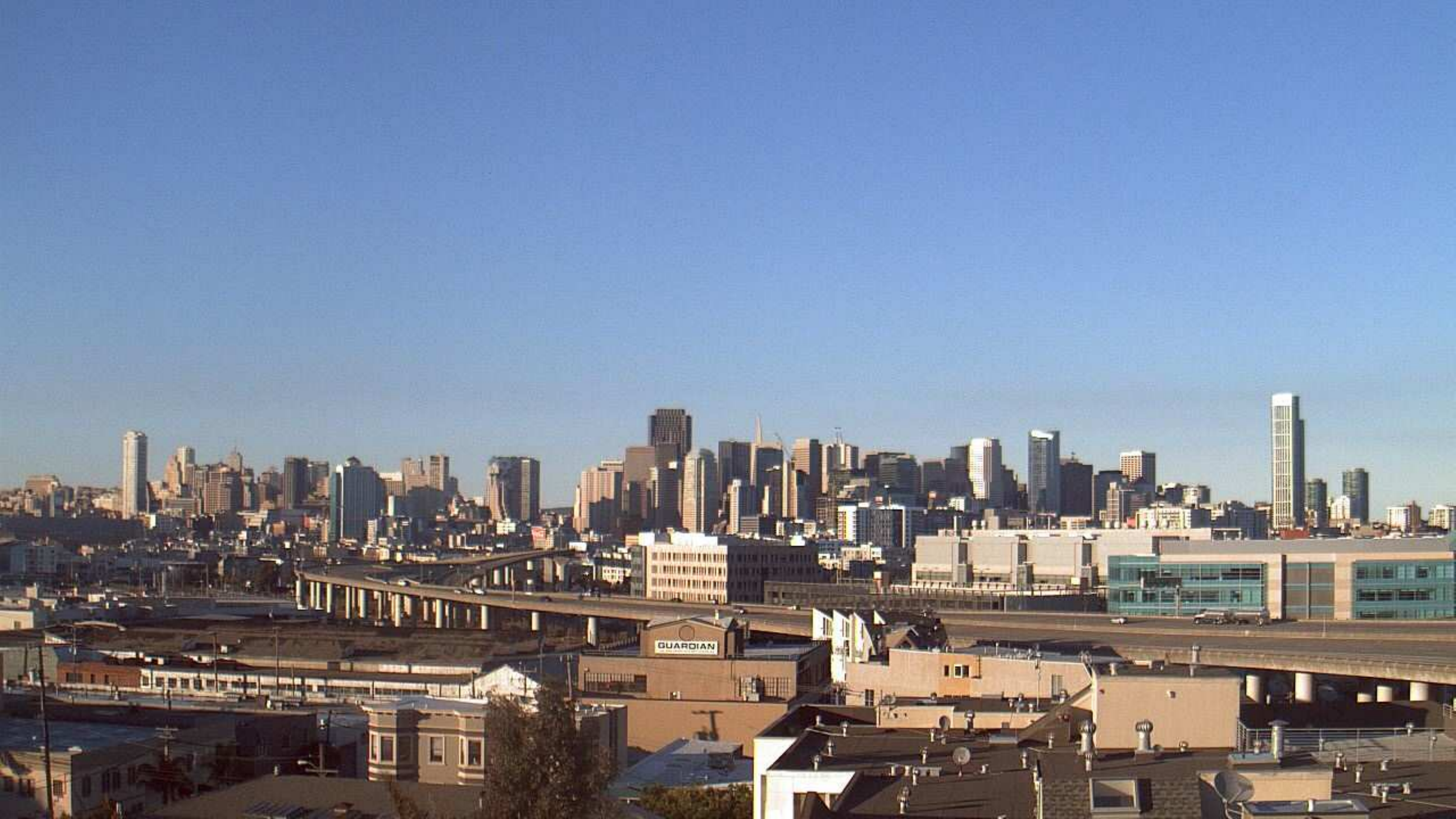} 
 \includegraphics[width=4cm]{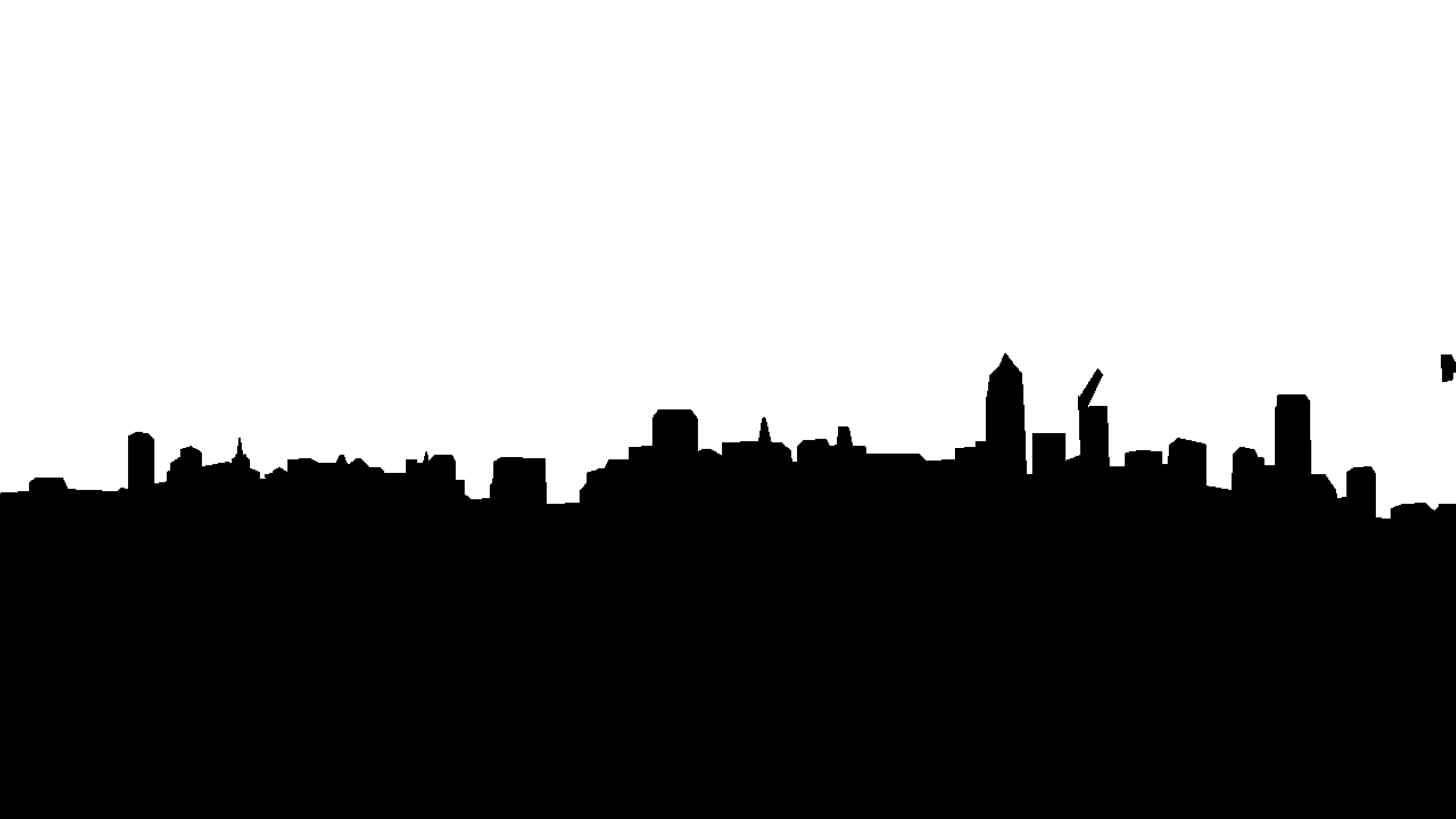} \\
 \includegraphics[width=4cm]{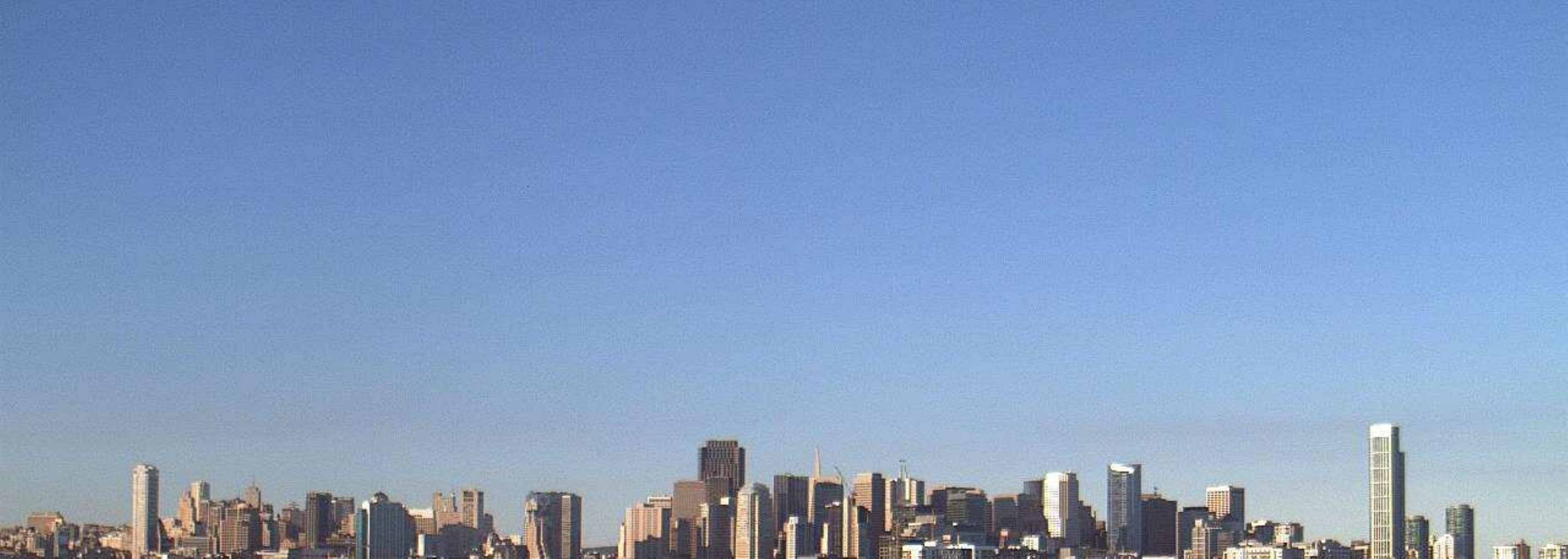} 
 \includegraphics[width=4cm]{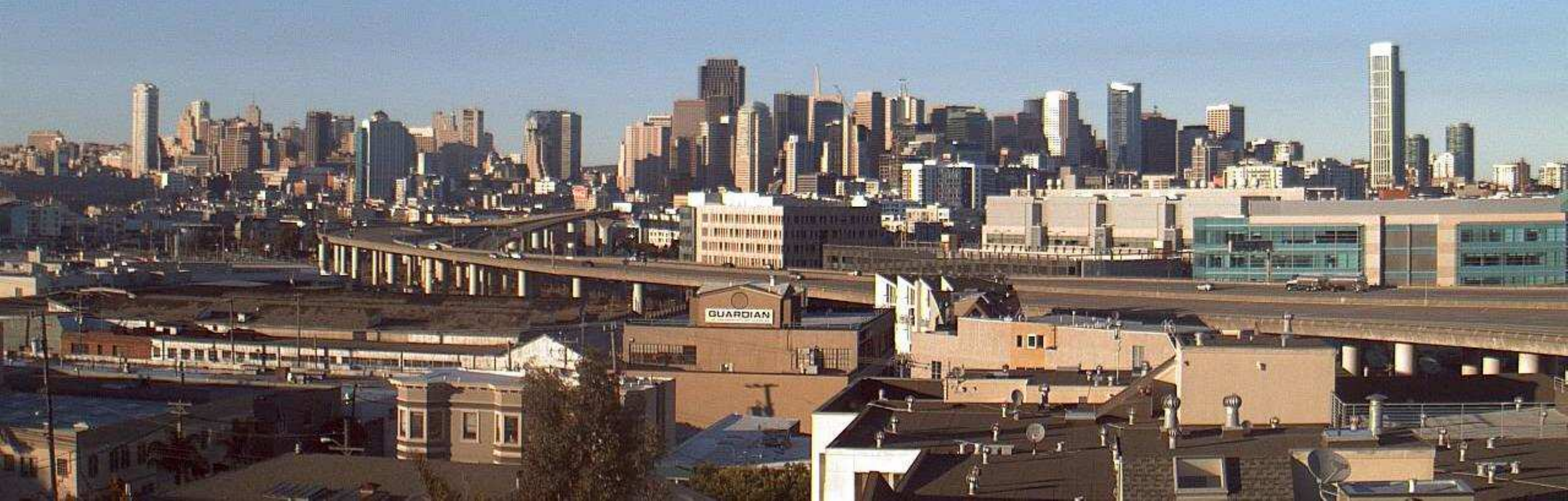} 
\caption{Top row: one sample scene image (left), and the corresponding mask of the sky region (right). Bottom row: the minimum bounding box of the sky region (left), and the minimum bounding box of the ground region (right).}
\label{fig:skyground}
\end{figure}

\textbf{Sky only vs. ground only.} The results shown in Table~\ref{tb:rnnperf_com} are obtained when visual information is extracted from the entire image. Intuitively, one may think that the sky region provides more information in estimating weather properties. This is why the works \cite{lu17} and \cite{chu17} extracted cloud features from the sky region to do weather condition classification. On the other hand, the first two rows of Fig.~\ref{fig:glasner} show that ground appearance sometimes provides important clues regarding temperature, e.g., whenever there is snow on the ground, it is colder. To quantify the influences of sky and ground regions on temperature estimation, first we manually segment the sky region for each scene. Then, the minimum bounding box of the sky region is fed to the proposed bidirectional LSTM model to obtain the performance yielded by sky-only information. We also feed the minimum bounding box of the ground region to obtain corresponding performance. Fig.~\ref{fig:skyground} shows samples of one scene, where the top row shows the original scene image and the corresponding sky mask, and the bottom row shows the minimum bounding boxes of the sky region and the ground region, respectively. Notice that because the sky region is irregular, the minimum bounding box may include parts of the ground, and vice versa. 

Table~\ref{tb:skyground} shows performance variations when visual information is extracted from different regions of the images in image sequences. Comparing the performance yielded by sky only with ground only, we found that in fact the ground region provides more information for temperature estimation. Visual variations in the sky region may just represent texture of cloud and intensity, and provide relatively fewer clues for estimating temperature. According to this experiment, we see that the best performance can be obtained when the entire image is considered. This conforms to the setting of previous works \cite{glasner15} and \cite{zhou17}. 

\begin{table}
\centering
\caption{Root mean square errors when visual information is extracted from different regions. } 
\begin{tabular}{c |c |c |c }
\hline
 & Sky only & Ground only & Entire Image \\
\hline
Avg. RMSE & 3.42 & 2.93 & \textbf{2.82}  \\
\hline
\end{tabular}
\label{tb:skyground}
\end{table}

\textbf{Variations of different daytime hours.}
As mentioned in Sec.~\ref{sec:implement}, only the image sequences captured at 11am are tested to obtain performance shown above. The main reason to use such setting is to align the settings of \cite{glasner15} and \cite{zhou17} to make a fair comparison. Besides, we are interested in whether image sequences captured in different daytime hours would yield performance variations. To show this, we respectively use image sequences at $h$ as the testing data, and the remaining is used for training. The hour $h$ is from 8am to 17pm.  

\begin{figure}
\centering
\includegraphics[width=8cm]{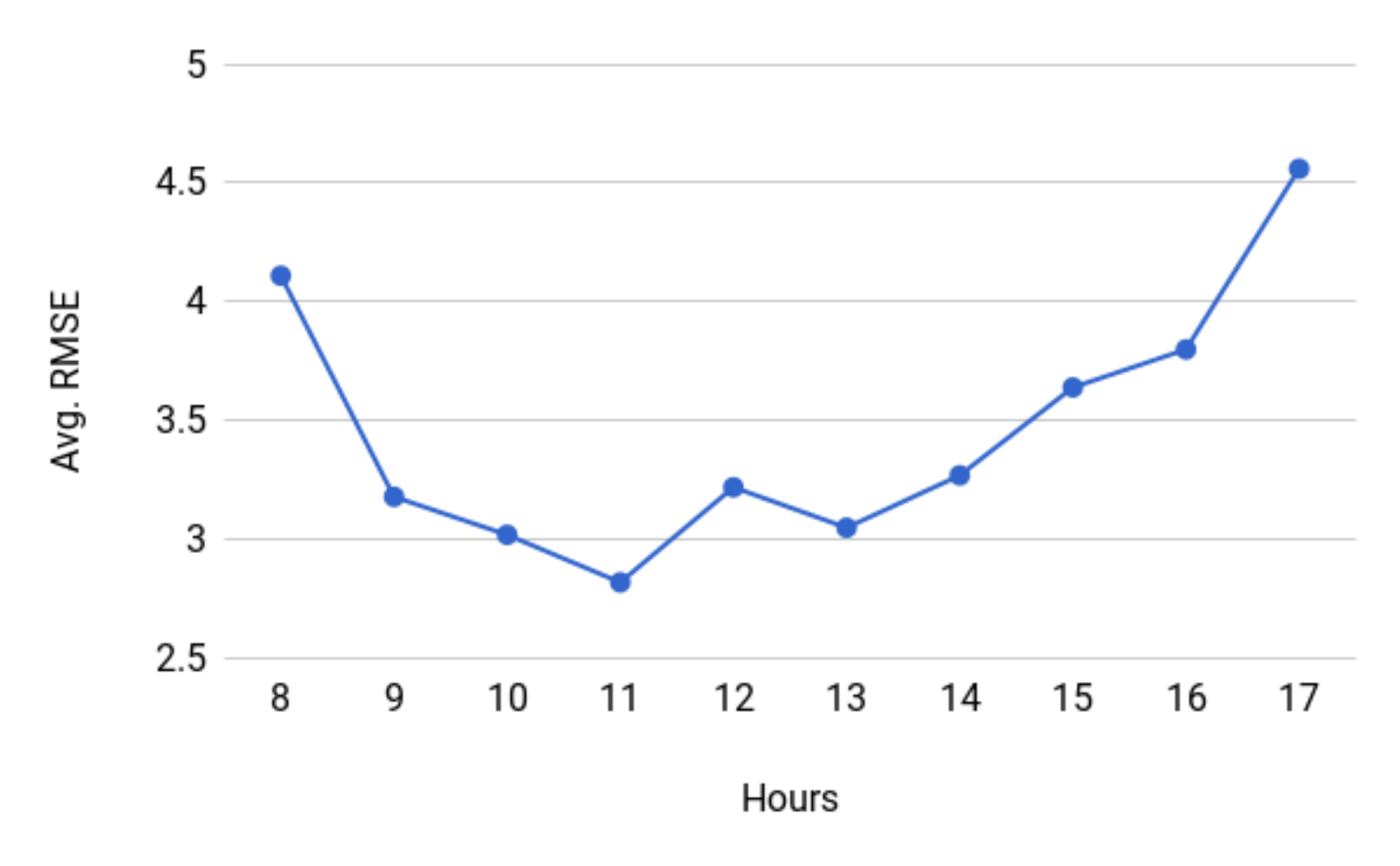} 
\caption{Variations of average RMSEs for image sequences captured in different daytime hours. }
\label{fig:RMSE_hour}
\end{figure}

Fig.~\ref{fig:RMSE_hour} shows variations of average RMSEs for image sequences captured in different daytime hours. Interestingly, the variation is significant when images captured in different daytime hours are processed. The best performance is obtained for images captured at 11 pm, which conforms to the selection of \cite{glasner15} and \cite{zhou17}. On the other hand, estimation errors at 8pm and 17am are much larger than others. This may be because the sunlight is maximal around the noon, and more robust visual information can be extracted. 

\begin{figure}
\centering
\includegraphics[width=8cm]{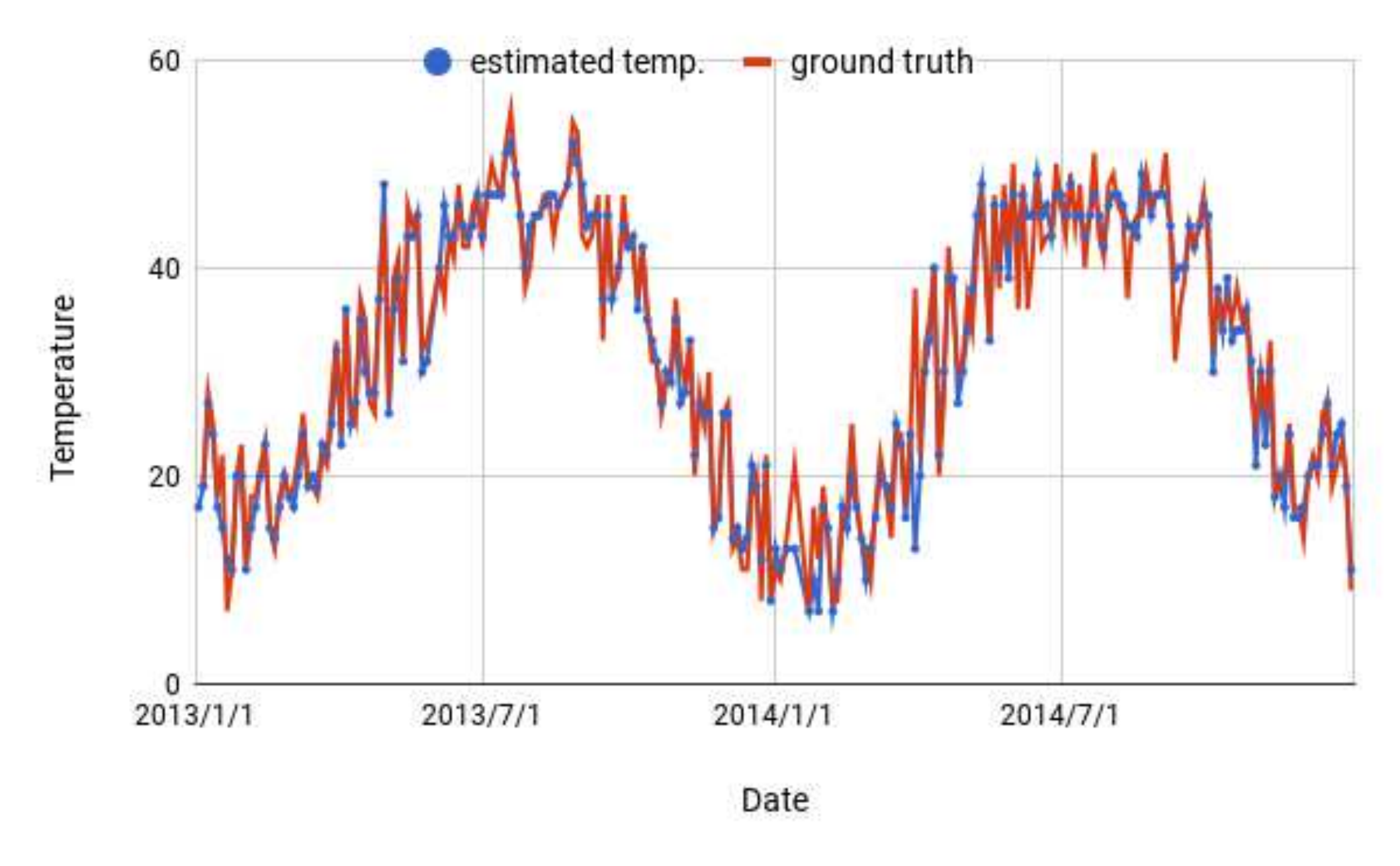}
\caption{Ground truth temperatures and the estimated temperatures of images captured at 11am by the camera (f) over two years.}
\label{fig:estvstruth}
\end{figure}

\textbf{Sample predictions.}
Fig.~\ref{fig:estvstruth} shows ground truth temperatures (red curve) and the estimated temperatures (blue curve) of images captured at 11am by the camera (f) mentioned in Table~\ref{tb:rnnperf_com} over two years (Jan. 2013 to Dec. 2014). Basically, we see that two curves correlate well with each other, conforming to the promising performance shown in Table~\ref{tb:rnnperf_com}. We also clearly observe the trend of climate change, i.e., higher temperature in July and August, and lower temperature in January and February.  

\begin{figure}
\centering
\includegraphics[width=4cm, height=3cm]{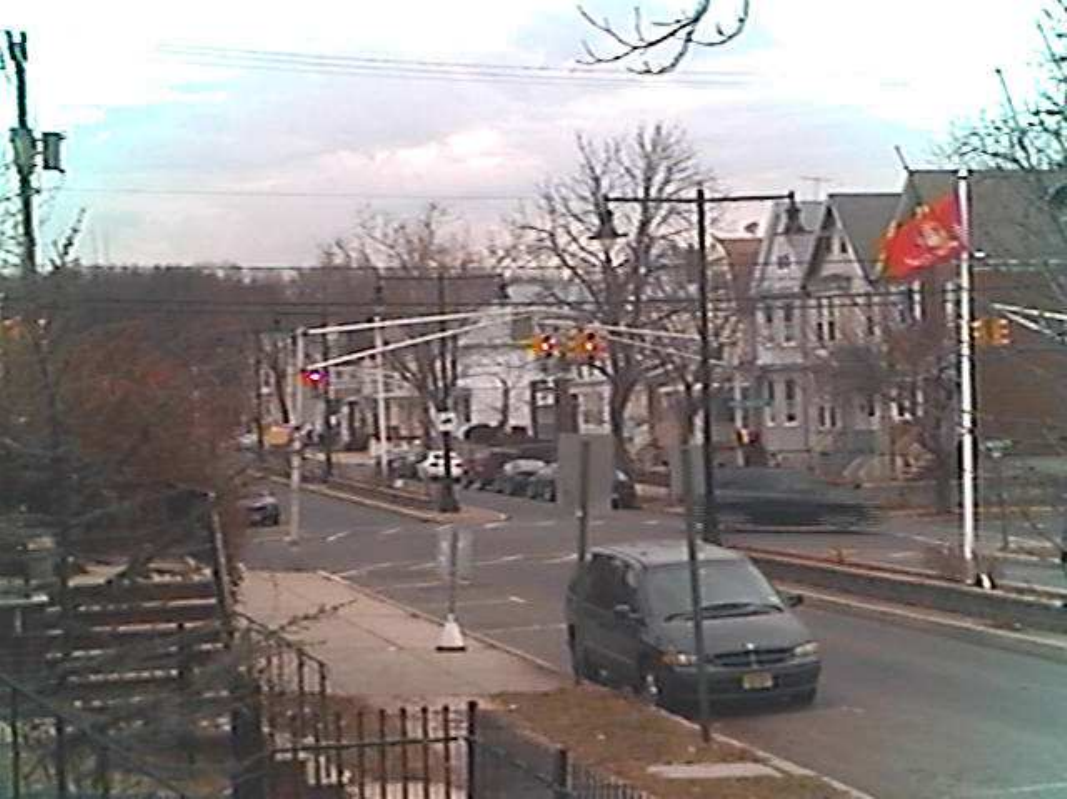}
\includegraphics[width=4cm, height=3cm]{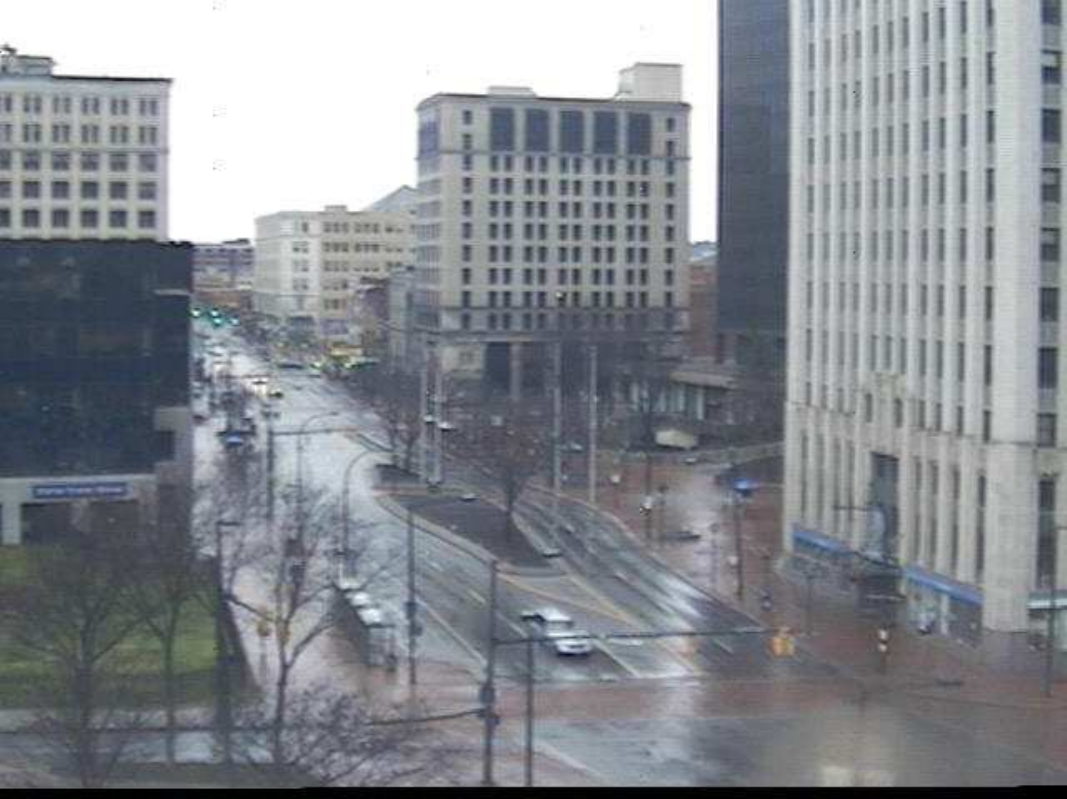}
\caption{Sample images of \enquote{successful cases}. For the left image, the truth temperature is $19^{\circ}$C, and the proposed bidirectional LSTM model estimates it as $19^{\circ}$C. For the right image, the truth temperature and the estimated one are $10^{\circ}$C and $11^{\circ}$C, respectively.}
\label{fig:case1}
\end{figure}

\begin{figure}
\centering
\includegraphics[width=4cm, height=3cm]{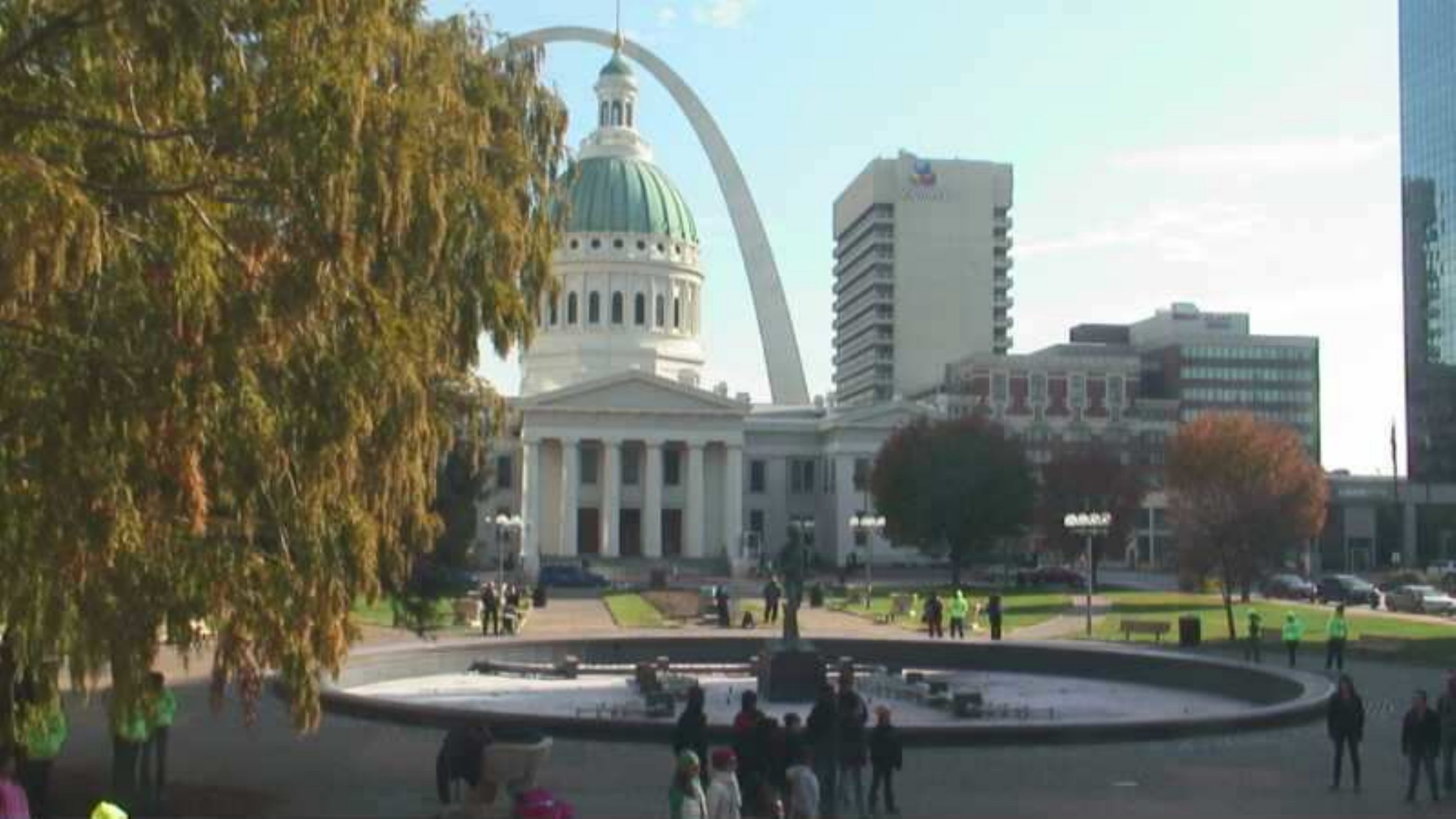}
\includegraphics[width=4cm, height=3cm]{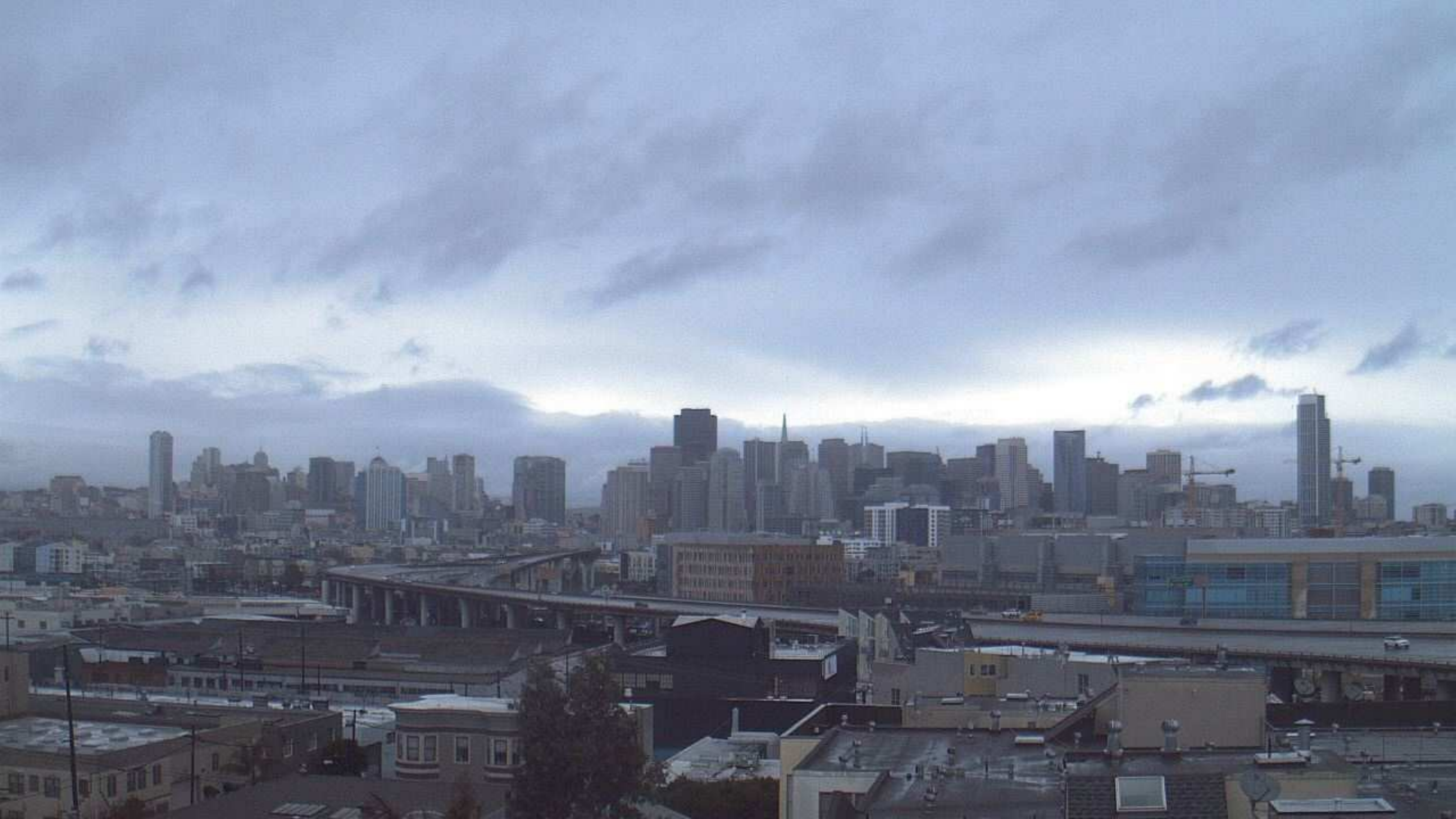}
\caption{Samples image of \enquote{failure cases}. For the left image, the truth temperature and the estimated one are $21^{\circ}$C and $38^{\circ}$C, respectively. For the right image, the truth temperature and the estimated one are $13^{\circ}$C and $21^{\circ}$C, respectively. }
\label{fig:case2}
\end{figure}

Fig.~\ref{fig:case1} and Fig.~\ref{fig:case2} show \enquote{successful estimation cases} and \enquote{failure estimation cases}, respectively. Looking only at one image may not allow us to figure out why temperature of some images are easier to estimate, and some are not. After a deeper inspection, we found that usually the failure cases occur when the temperature values of consecutive days fluctuate significantly. This is expectable. We also found some ground truths are obviously incorrect, e.g., a snowy scene has the truth temperature of $42^{\circ}$C. These errors need to be fixed manually in the future to elaborate more robust performance evaluation.   

It is interesting to know what visual clues are more important in temperature prediction. To get some insights for this issue, we take all images captured closest to 11am by a camera, and divide each image into non-overlapping $5 \times 5$ block. Based on pixels' RGB colors in blocks corresponding to the same position, e.g., the $i$th block in images captured on the first day, the second day, and so on, we calculate the standard deviations of R, G, and B, and then average these three standard deviations as a value $\rho$. The value $\rho$ is then normalized into the range from 0 to 255 by $\hat{\rho} = \dfrac{\rho - \min_{\rho}}{\max_{\rho} - \min_{\rho}} \times 255$, where $\max_{\rho}$ and $\min_{\rho}$ are the maximum $\rho$'s and minimum $\rho$'s of all blocks. Finally, we visualize the value $\hat{\rho}$ of each block. A block with a larger $\hat{\rho}$ value indicates higher color variations on different days, and probably conveys more information in predicting temperature. 

\begin{figure}
\centering
\includegraphics[width=4cm, height=3cm]{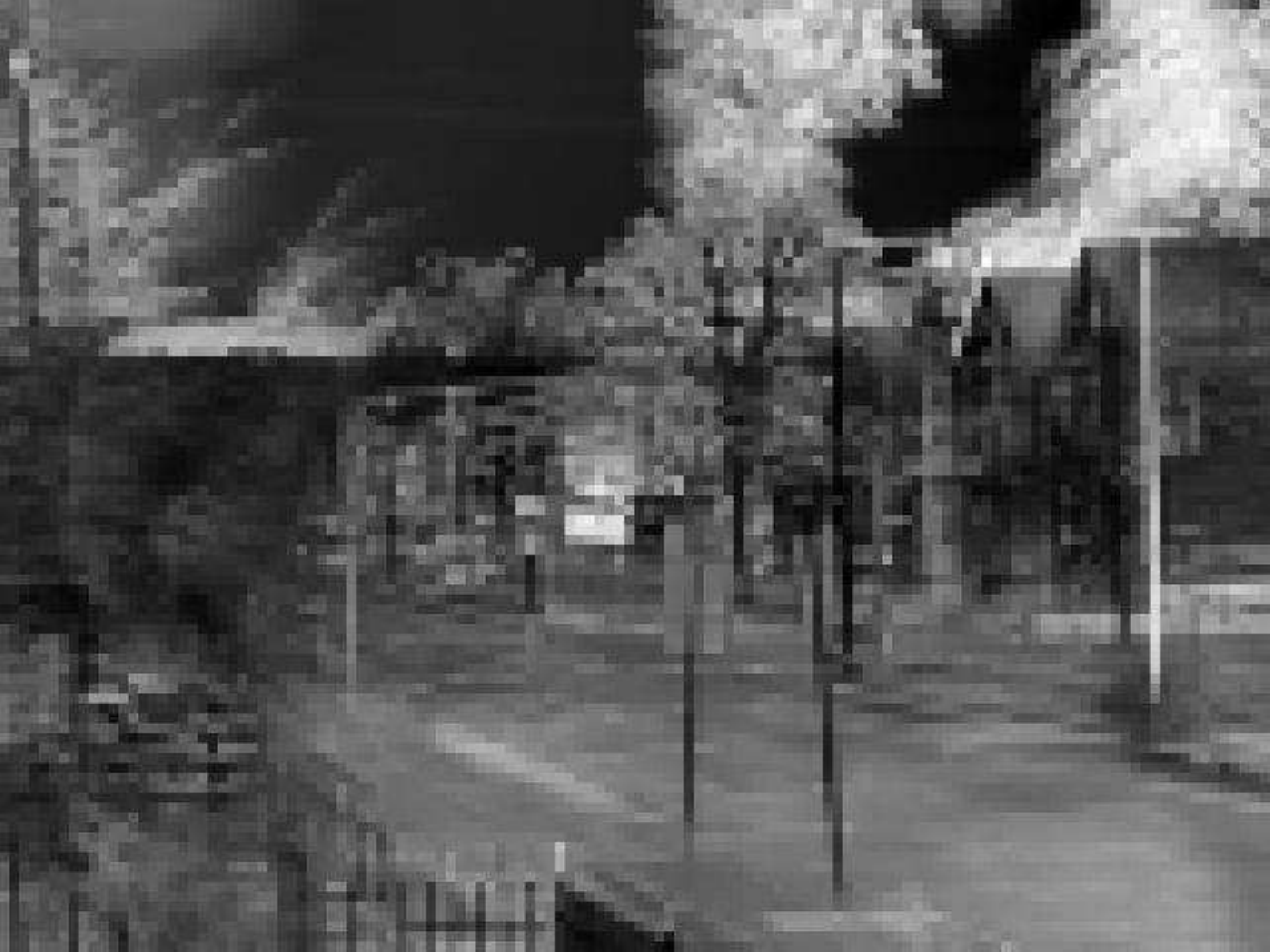}
\includegraphics[width=4cm, height=3cm]{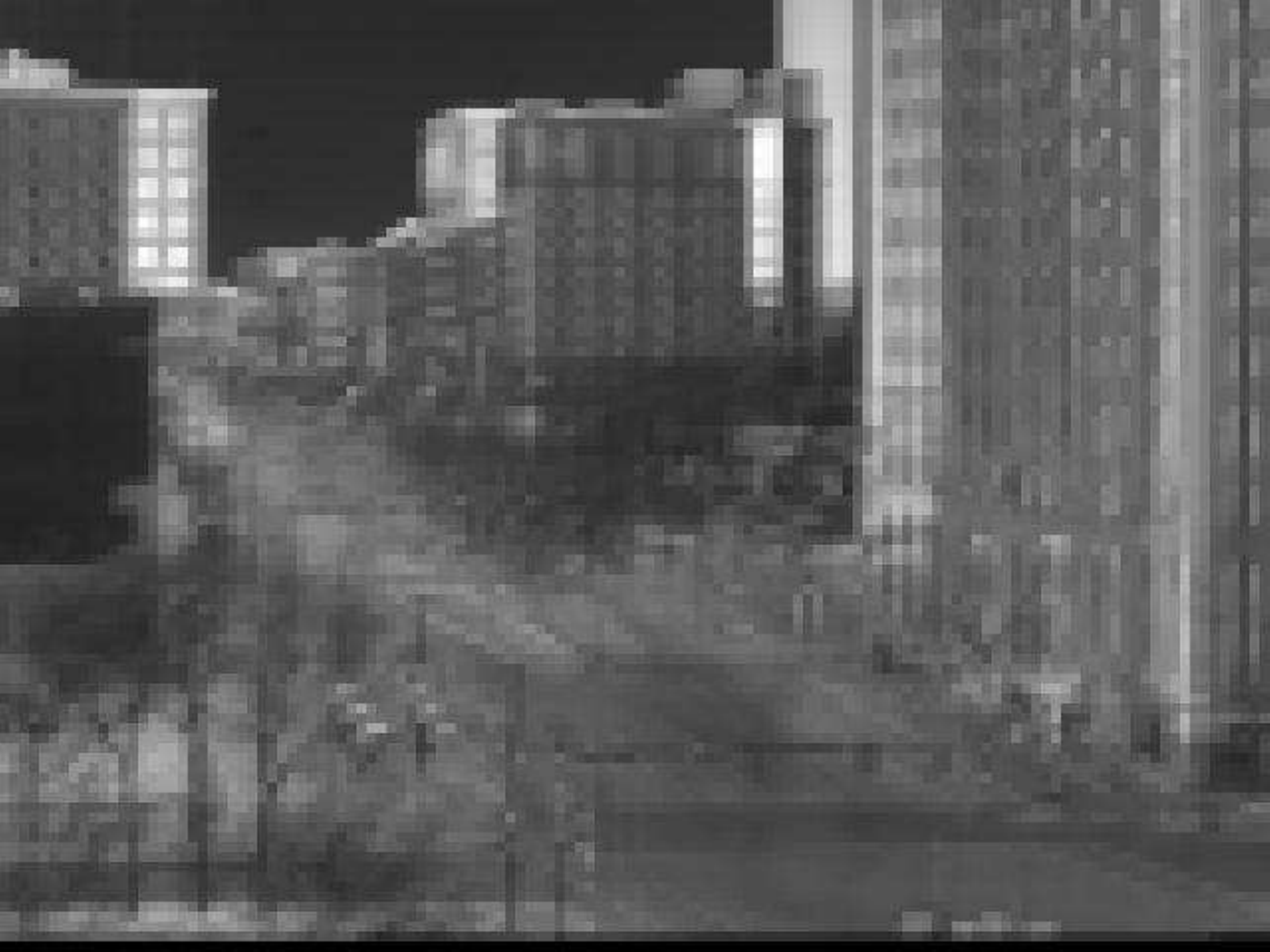}
\caption{Saliency maps showing the importance of different pixels on temperature prediction. }
\label{fig:saliency}
\end{figure}

Fig.~\ref{fig:saliency} shows the visualization results corresponding to the scenes shown in Fig.~\ref{fig:case1}. We see that, for the first scene, blocks on tree leaves are more salient, and may be more informative in temperature prediction. For the second scene, blocks on some parts of building facades are more salient. These results are similar to the discussion mentioned in \cite{glasner15}, and could be interesting clues for future studies. 

%-------------------------------------------------------------------------

\section{Conclusion and Discussion}
In this work, we present deep models to estimate ambient temperature of a single image or the last image in an image sequence. For the first task, we verify that the CNN-based approach is promising, and when more training data is available better performance can be obtained. For the second task, we consider temporal evolution of visual appearance, and propose an RNN-based approach to \enquote{forecast} the temperature of the last image. State-of-the-art performance can be obtained by the proposed method. We also discuss how performance varies when information is just extracted from the sky region or from the ground region, and how performance varies when images captured in different daytime hours are processed. 

In the future, more weather properties like humidity and wind speed may be estimated by the proposed methods. Furthermore, we can investigate which region in a scene provides more clues in temperature estimation, based on the currently emerged attention networks. Exploring the relationship between weather properties and vision would be interesting in a wide range of works and applications. 

\section*{Acknowledgement}
This work was partially supported by the Ministry of Science and Technology of Taiwan under the grant MOST 105-2628-E-194-001-MY2 and MOST 106-3114-E-002-009.

{\small
\bibliographystyle{ieee}
\bibliography{weather}
}

\end{document}